\PassOptionsToPackage{usenames,dvipsnames,table,hyperref}{xcolor}
\documentclass[11pt,a4paper]{article}
\usepackage{emnlp-ijcnlp-2019}
\usepackage{times}
\usepackage{latexsym}
\usepackage{url}
\usepackage{amssymb}
\usepackage{graphicx}
\usepackage{times}
\usepackage{helvet}
\usepackage{courier}
\usepackage{balance}  
\usepackage{mdwlist}
\usepackage{graphics}
\usepackage{color}
\usepackage{rotating}
\usepackage{booktabs}
\usepackage{epsfig}
\usepackage{alltt}
\usepackage{moreverb}
\usepackage{fancyvrb}
\usepackage{enumerate}
\usepackage{xspace}
\usepackage{relsize}
\usepackage{amsmath}
\usepackage{amsfonts}
\usepackage{txfonts}
\usepackage{tabularx}
\usepackage{rotating}
\usepackage{multirow}
\usepackage{booktabs}
\usepackage{float}
\usepackage{array,etoolbox}
\usepackage{enumitem}
\usepackage{subfig}
\usepackage{arydshln}
\usepackage{setspace}
\usepackage{rotating}
\usepackage{pbox}
\usepackage{todonotes}
\usepackage{bm}
\usepackage{algorithm}
\usepackage[noend]{algpseudocode}
\usepackage{tikz}
\def\checkmark{\tikz\fill[scale=0.4](0,.35) -- (.25,0) -- (1,.7) -- (.25,.15) -- cycle;} 
\usepackage{ wasysym }
\usepackage[export]{adjustbox}
\newsavebox{\measurebox}
\usepackage{caption}
\usepackage{etoolbox}
\usepackage[toc,page]{appendix}

\aclfinalcopy 

\newtoggle{draft}
\togglefalse{draft}
\iftoggle{draft}{
\newcommand{\dk}[1]{\textcolor{Cyan}{[#1 \textsc{--dk}]}}
\newcommand{\taehee}[1]{\textcolor{Red}{[#1 \textsc{--taehee}]}}
\newcommand{\lucas} [1]{\textcolor{Orange}{[#1 \textsc{--lucas}]}}
\newcommand{\ed}[1]{\textcolor{Green}{[#1 \textsc{--ed}]}}
\newcommand{\fillthis}[1]{\textcolor{Yellow}{[#1]}}

}{
\newcommand{\dk}[1]{}
\newcommand{\taehee}[1]{}
\newcommand{\lucas}[1]{}
\newcommand{\ed}[1]{}
\newcommand{\fillthis}[1]{}

}

\newcommand{\V}{\ensuremath{\mathcal{V}}\xspace}
\newcommand{\Search}{\ensuremath{\mathcal{S}}\xspace}

\algrenewcommand\algorithmicforall{\textbf{foreach}}

\newcommand{\ignore}[1]{}   
\newcolumntype{Y}{>{\centering\arraybackslash}X}

\begin{document}

\title{Earlier Isn't Always Better: \\Sub-aspect 
Analysis on Corpus and System Biases in Summarization}

\author{
Taehee Jung\thanks{\@\@\,\, Equal contribution, name order decided by coin flip.}\, $^\heartsuit$ \@ Dongyeop Kang\footnotemark[1]\, $^\clubsuit$ \quad \quad Lucas Mentch $^\heartsuit$ \quad \quad Eduard Hovy $^\clubsuit$\\
$^\heartsuit$Department of Statistics, University of Pittsburgh, Pittsburgh, PA, USA\\
$^\clubsuit$School of Computer Science, Carnegie Mellon University, Pittsburgh, PA, USA \\
{\tt $\{$taj41,lkm31$\}$@pitt.edu}\quad {\tt $\{$dongyeok,hovy$\}$@cs.cmu.edu   }}

\maketitle

\begin{abstract}
Despite the recent developments on neural summarization systems, the underlying logic behind the improvements from the systems and its corpus-dependency remains largely unexplored.  Position of sentences in the original text, for example, is a well known bias for news summarization. Following in the spirit of the claim that summarization is a combination of sub-functions, we define three sub-aspects of summarization:  \texttt{position}, \texttt{importance}, and \texttt{diversity} and conduct an extensive analysis of the biases of each sub-aspect with respect to the domain of nine different summarization corpora (e.g., news, academic papers, meeting minutes, movie script, books, posts).  We find that while \texttt{position} exhibits substantial bias in news articles, this is not the case, for example, with academic papers and meeting minutes. 
Furthermore, our empirical study shows that different types of summarization systems (e.g., neural-based) are composed of different degrees of the sub-aspects.
Our study provides useful lessons regarding consideration of underlying sub-aspects when collecting a new summarization dataset or developing a new system.
\end{abstract}

\section{Introduction}
Despite numerous recent developments in neural summarization systems \cite{narayan2018ranking,nallapatiabstractive,see2017get,kedzie2018content,gehrmann2018bottom,paulus2017deep} the underlying rationales behind the improvements and their dependence on the training corpus remain largely unexplored.
\citet{edmundson1969new} put forth the position hypothesis: important sentences appear in preferred positions in the document.
\citet{lin1997identifying} provide a method to empirically identify such positions.
Later, \citet{hong2014improving} showed an intentional lead bias in news writing, suggesting that sentences appearing early in news articles are more important for summarization tasks.
More generally, it is well known that recent state-of-the-art models \cite{nallapatiabstractive,see2017get} are often marginally better than the first-k baseline on single-document news summarization.

In order to address the position bias of news articles, \citet{narayan2018don} collected a new dataset called XSum to create single sentence summaries that include material from multiple positions in the source document.
\citet{kedzie2018content} showed that the position bias in news articles is not the same across other domains such as meeting minutes \cite{carletta2005ami}.

In addition to \texttt{position}, \citet{lin2012learning} defined other sub-aspect functions of summarization including \texttt{coverage}, \texttt{diversity}, and \texttt{information}. 
\citet{lin2011class} claim that many existing summarization systems are instances of mixtures of such sub-aspect functions; for example, maximum marginal relevance (MMR) \cite{carbonell1998use} can be seen as an combination of diversity and importance functions. 

Following the sub-aspect theory, we explore three important aspects of summarization (\S\ref{sec:multi_aspects}):
\textsc{position} for choosing sentences by their position, \textsc{importance} for choosing relevant contents, and \textsc{diversity} for ensuring minimal redundancy between summary sentences. 

We then conduct an in-depth analysis of these aspects over nine different domains of summarization corpora (\S\ref{sec:data}) including news articles, meeting minutes, books, movie scripts, academic papers, and personal posts.
For each corpus, we investigate which aspects are most important and develop a notion of \textbf{corpus bias} (\S\ref{sec:analysis_corpora}). We provide an empirical result showing how current summarization systems are compounded of which sub-aspect factors called \textbf{system bias} (\S\ref{sec:analysis_system}).
At last, we summarize our actionable messages for future summarization researches (\S\ref{sec:conclusion}).
We summarize some notable findings as follows:

\begin{figure}[t]
\vspace{0mm}
\centering
\sbox{\measurebox}{%
\begin{minipage}[b]{.8\linewidth}
\includegraphics[trim=0.9cm 0.4cm 0.7cm 12cm,clip,width=.99\linewidth]{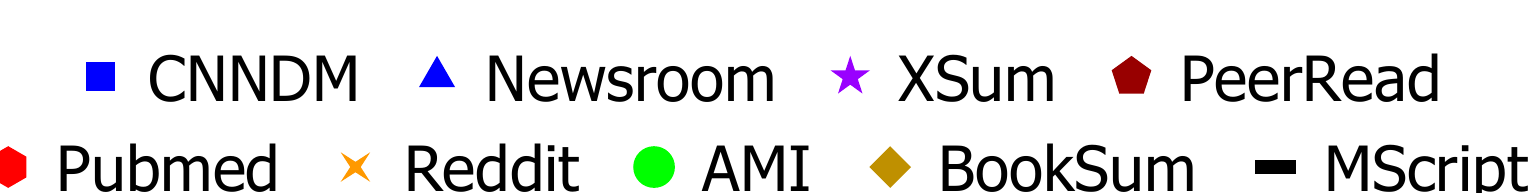}
\subfloat[\footnotesize{Corpus bias}]{
\includegraphics[trim=3cm 4.2cm 3cm 1.05cm,clip,width=.99\linewidth]{figs/triangle/triangle_all_corpus.pdf}
}
\end{minipage}}
\usebox{\measurebox}
\begin{minipage}[b][\ht\measurebox][s]{.8\linewidth}
\includegraphics[trim=0.7cm 13.5cm 0.6cm 0.4cm,clip,width=.99\linewidth]{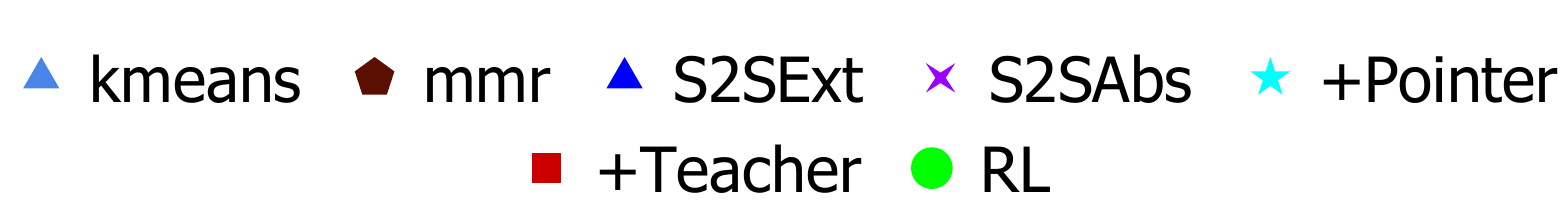}
\subfloat[\footnotesize{System bias}]{
\includegraphics[trim=2cm 2.6cm 2cm 2.8cm,clip,width=.99\linewidth]{figs/triangle/triangle_sys_cnndm.pdf}

}
\end{minipage}
\caption{\label{fig:triangle_all} Corpus and system biases with the three sub-aspects, showing what portion of aspect is used for each corpus and each system. 
The portion is measured by calculating ROUGE score  between (a) summaries obtained from each aspect and target summaries or (b) summaries obtained from each aspect and each system.}
\vspace{0mm}
\end{figure}

\begin{itemize}[noitemsep,topsep=0pt,leftmargin=*]
\item Summarization of personal post and news articles except for XSum \cite{narayan2018don} are biased to the position aspect, while academic papers are well balanced among the three aspects (see Figure \ref{fig:triangle_all} (a)).
Summarizing long documents (e.g.\ books and movie scripts) and conversations (e.g.\ meeting minutes) are extremely difficult tasks that require multiples aspects together.
\item Biases do exist in current summarization systems (Figure \ref{fig:triangle_all} (b)). Simple ensembling of multiple aspects of systems show comparable performance with simple single-aspect systems.  
\item Reference summaries in current corpora include less than 15\% of new words that do not appear in the source document, except for abstract text of academic papers.
\item Semantic volume \cite{yogatama2015extractive} overlap between the reference and model summaries is not correlated with the hard evaluation metrics such as ROUGE \cite{lin2004rouge}.
\end{itemize}



\section{Related Work}\label{sec:related}
We provide here a brief review of  prior work on summarization biases. 
\citet{lin1997identifying} studied the position hypothesis, especially in the news article writing \cite{hong2014improving,narayan2018don} but not in other domains such as conversations \cite{kedzie2018content}. 
\citet{narayan2018don} collected a new corpus to address the bias by compressing multiple contents of source document in the single target summary.
In the bias analysis of systems, \citet{lin2012learning,lin2011class} studied the sub-aspect hypothesis of summarization systems.
Our study extends the hypothesis to various corpora as well as systems.
With a specific focus on \texttt{importance} aspect, a recent work \cite{peyrard-2019-simple} divided it into three sub-categories; redundancy, relevance, and informativeness, and provided quantities of each to measure.
Compared to this, ours provide broader scale of sub-aspect analysis across various corpora and systems.

We analyze the sub-aspects on different domains of summarization corpora:
news articles \cite{nallapatiabstractive,N18-1065,narayan2018don}, academic papers or journals \cite{kang18naacl,kedzie2018content}, movie scripts \cite{gorinski2015movie}, books \cite{mihalcea2007explorations}, personal posts \cite{ouyang2017crowd}, and meeting minutes \cite{carletta2005ami} as described further in \S\ref{sec:data}.

Beyond the corpora themselves, a variety of summarization systems have been developed:
\citet{mihalcea2004textrank,erkan2004lexrank} used graph-based keyword ranking algorithms.
\citet{lin2010multi,carbonell1998use} found summary sentences which are highly relevant but less redundant.
\citet{yogatama2015extractive} used semantic volumes of bigram features for extractive summarization.
Internal structures of documents have been used in summarization: 
syntactic parse trees~\cite{woodsend2011learning,cohn2008sentence},
topics~\cite{zajic2004bbn,lin2000automated},
semantic word graphs~\cite{mehdad2014abstractive,gerani2014abstractive,ganesan2010opinosis,filippova2010multi,boudin2013keyphrase}, and abstract meaning representation~\cite{liu2015toward}.
Concept-based Integer-Linear Programming (ILP) solver~\cite{mcdonald2007study} is used for optimizing the summarization problem~\cite{gillick2009scalable,banerjeemulti2015ijcai,boudin2015concept,berg2011jointly}.
\citet{durrett2016learning} optimized the problem with grammatical and anarphorcity constraints.

With a large scale of corpora for training, neural network based systems have recently been developed. 
In abstractive systems, \citet{rush2015neural} proposed a local attention-based sequence-to-sequence model.
On top of the seq2seq framework, many other variants have been studied using convolutional networks \cite{cheng2016neural,allamanis2016convolutional}, pointer networks \cite{see2017get}, scheduled sampling \cite{bengio2015scheduled}, and reinforcement learning \cite{paulus2017deep}. In extractive systems, different types of encoders \cite{cheng2016neural,nallapati2017summarunner,kedzie2018content} and optimization techniques \cite{narayan2018ranking} have been developed.
Our goal is to explore which types of systems learns which sub-aspect of summarization.

\section{Sub-aspects of Summarization}\label{sec:multi_aspects}

We focus on three crucial aspects
: \textsc{Position}, \textsc{Diversity}, and \textsc{Importance}.
For each aspect, we use different extractive algorithms to \textbf{capture how much of the aspect is used in the oracle extractive summaries}\footnote{See \S\ref{sec:metrics} for our oracle set construction.}.
For each algorithm, the goal is to select $k$ extractive summary sentences (equal to the number of sentences in the target summaries for each sample) out of $N$ sentences appearing in the original source. 
The chosen sentences or their indices will be used to calculate the various evaluation metrics described in \S\ref{sec:metrics}

For some algorithms below, we use vector representation of sentences.
We parse a document $x$ into a sequence of sentences $x = x_1..x_N$ where each sentence consists of a sequence of words $x_i = w_1..w_s$. 
Each sentence is then encoded:
\begin{equation}\label{eq:bert}
\textsc{E}(x_i) = \texttt{BERT} (w_{i,1}..w_{i,s})
\end{equation}
where \texttt{BERT} \cite{devlin2018bert} is a pre-trained bidirectional encoder from transformers \cite{vaswani2017attention}\footnote{The other encoders such as averaging word embeddings \cite{pennington2014glove} show comparable performance.}.
We use the last layer from \texttt{BERT} as a representation of each token, and then average them to get final representation of a sentence.
All tokens are lower cased. 

\subsection{\textsc{Position}}\label{sec:multi_aspects_position}
Position of sentences in the source has been suggested as a good indicator for choosing summary sentences, especially in news articles \cite{lin1997identifying,hong2014improving,see2017get}.
We compare three position-based algorithms: \textbf{First}, \textbf{Last}, and \textbf{Middle}, by simply choosing $k$ number of sentences in the source document from these positions.

\subsection{\textsc{Diversity}}\label{sec:multi_aspects_diversity}
\citet{yogatama2015extractive} assume that extractive summary sentences which maximize the semantic volume in a distributed semantic space are 
the most diverse but least redundant sentences.
Motivated by this notion, our goal is to find a set of $k$ sentences that maximizes the volume size of them in a continuous embedding space like the BERT representations in Eq \ref{eq:bert}. 
Our objective is to find the optimal search function \Search that maximizes the volume size  \V of searched sentences: $\arg\max_{1..k} \V\, (\Search_{1..c}\,(\textsc{E}(x_1),\ldots,\textsc{E}(x_N))) $.
\begin{figure}[h]
\vspace{-2mm}
    \centering
     {
      \subfloat[\footnotesize{Default}]{
     \includegraphics[trim=1cm 2cm 11cm 2cm,clip,width=.3\linewidth]{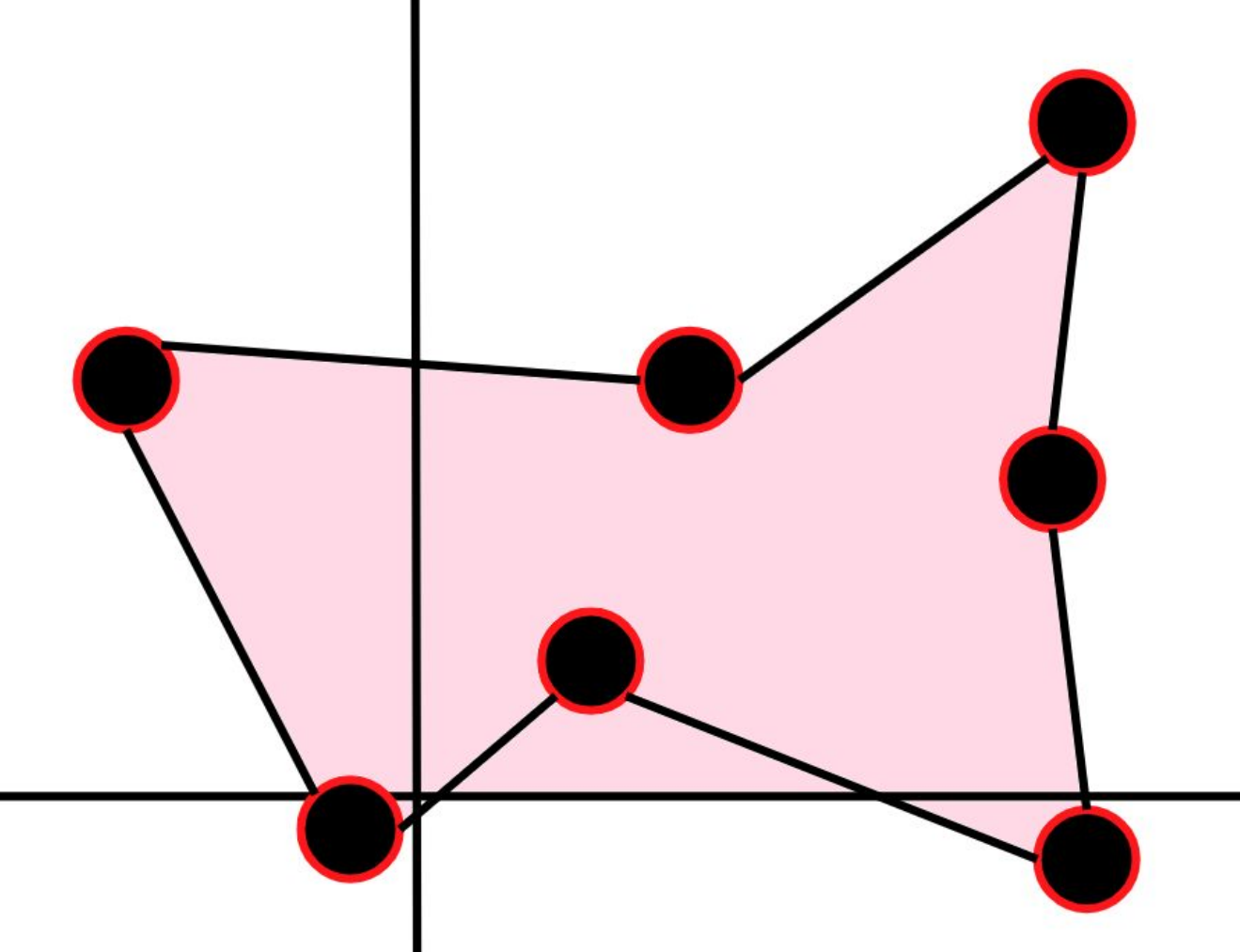}
     }
      \subfloat[\footnotesize{Heuristic}]{
     \includegraphics[trim=1cm 2cm 11cm 2cm,clip,width=.3\linewidth]{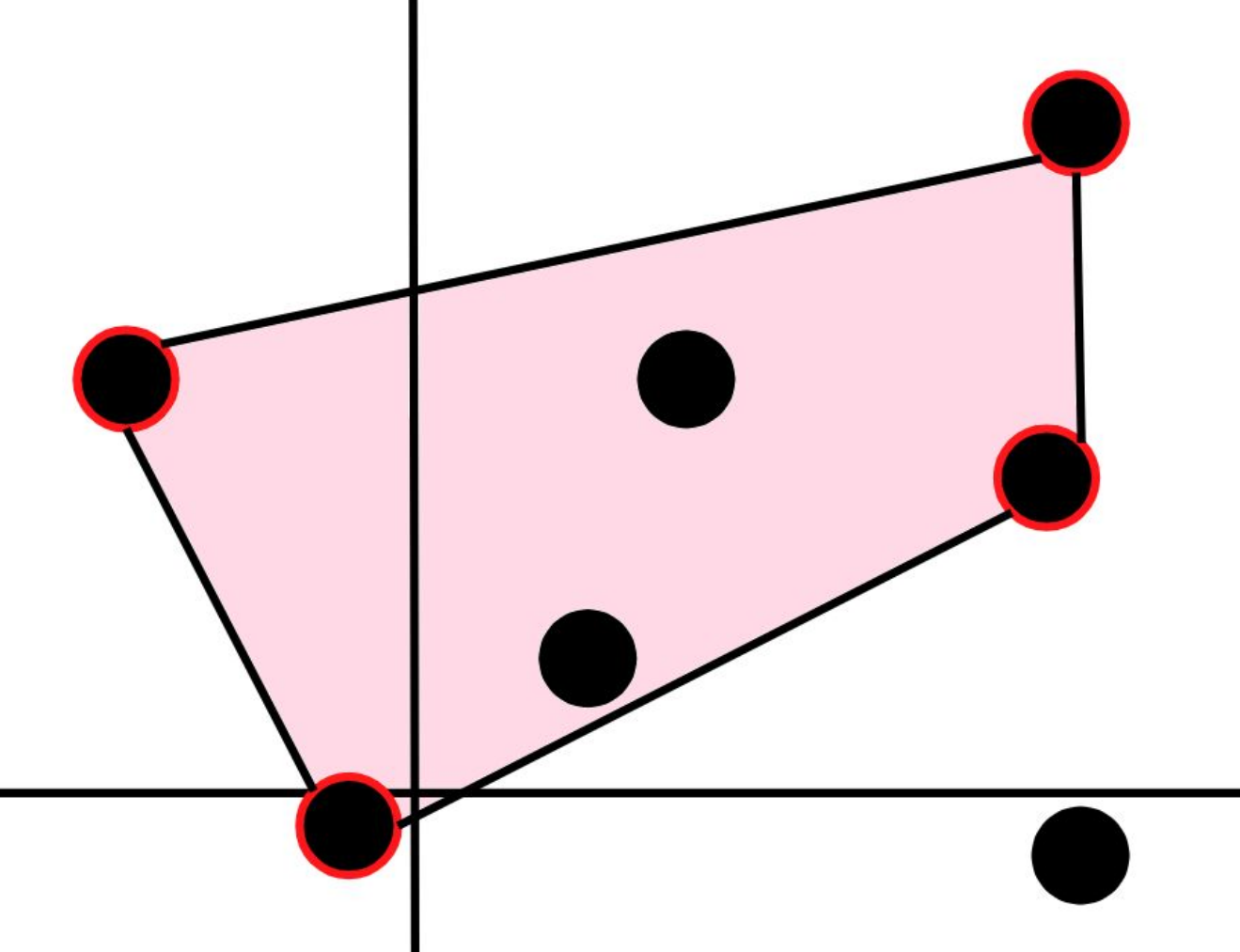}
     }
      \subfloat[\footnotesize{ConvexFall}]{
     \includegraphics[trim=1cm 2cm 11cm 2cm,clip,width=.3\linewidth]{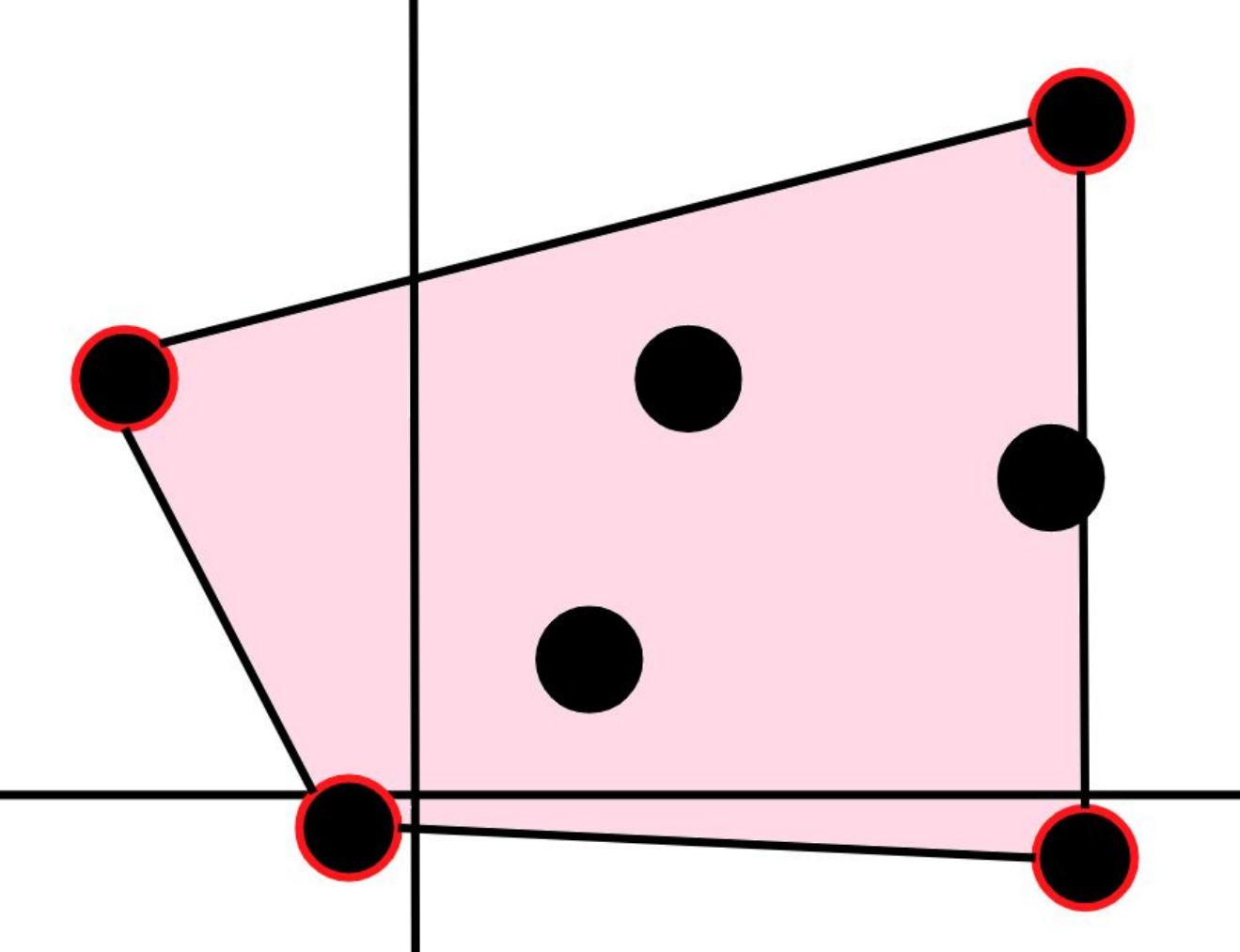}
     }     
     }
    \caption{\label{fig:volume_selection} Volume maximization functions. Black dots are sentences in source document, and red dots are chosen summary sentences. The red-shaded polygons are volume space of the summary sentences.
    }
    \vspace{-2mm}
\end{figure}

If $k$=$N$, we use every sentence from the source document.
(Figure \ref{fig:volume_selection} (a)).
However, its volume space does not guarantee to maximize the volume size because of the non-convex polygonality.
In order to find a convex maximum volume, we consider two different algorithms described below.

\textbf{Heuristic.}
\citet{yogatama2015extractive} heuristically choose a set of summary sentences using a greedy algorithm:
It first chooses a sentence which has the farthest vector representation from the centroid of whole source sentences, and then repeatedly finds sentences whose representation is farthest from the centroid of vector representations of the chosen sentences. Unlike the original algorithm in \cite{yogatama2015extractive} restricting the number of words, we constrain the total number of selected sentences to $k$.
This heuristic algorithm can fail to find the maximum volume depending on its starting point and/or the farther distance between two points detected (Figure \ref{fig:volume_selection} (b)).

\textbf{ConvexFall.}
Here we first find the convexhull\footnote{Definition: a set of points is defined as the smallest convex set that includes the points.} using Quickhull \cite{barber1996quickhull}, implemented by Qhull library\footnote{\url{http://www.qhull.org/}}.
It guarantees the maximum volume size of selected points with minimum number of points (Figure \ref{fig:volume_selection} (c)).
However, it does not reduce a redundancy between the points over the convex-hull, and usually choose larger number of sentences than $k$.  
\citet{marcu1999discourse} shows an interesting study regarding an importance of sentences: given a document, if one deletes the least central sentence from the source text, then at some point the similarity with the reference text rapidly drops at sudden called the \textit{waterfall} phenomena. 
Motivated by his study, we similarly prune redundant sentences from the set chosen by convex-hull search. 
For each turn, the sentence with the lowest volume reduction ratio is pruned until the number of remaining sentences is equivalent to $k$.

\subsection{\textsc{Importance}}\label{sec:multi_aspects_importance}
We assume that contents that repeatedly occur in one document contain \textit{important} information. 
We find sentences that are nearest to the neighbour sentences using two distance measures: 
\textbf{N-Nearest} calculates an averaged Pearson correlation between one and the rest for all source sentence vector representations. $k$ sentences having the highest averaged correlation are selected as final extractive summaries.  
On the other hand, \textbf{K-Nearest} chooses the $K$ nearest sentences per each sentence, and then averages distances between each nearest sentence and the selected one. The one has the lowest averaged distance is chosen. This calculation is repeated $k$ times and the selected sentences are removed from the remaining pool. 


\section{Metrics}\label{sec:metrics}

In order to determine the aspects most crucial to the summarization task, we use three evaluation metrics:

\textbf{\texttt{ROUGE}} is Recall-Oriented Understudy for Gisting Evaluation \cite{lin2000automated} for evaluating summarization systems.
We use ROUGE-1 (R1), ROUGE-2 (R2), and ROUGE-L (RL) F-measure scores which corresponds to uni-gram, bigrams and longest common subsequences, respectively, and their averaged score (R).

\textbf{\texttt{Volume Overlap (VO)} ratio.}
Hard metrics like ROUGE often ignore semantic similarities between sentences. 
Based on the volume assumption in \cite{yogatama2015extractive}, we measure overlap ratio of two semantic volumes calculated by the model and target summaries.
We obtain a set of vector representations of the reference summary sentences $\hat{Y}$ and the model summary sentences $Y$ predicted by any algorithm $algo$ in \S\ref{sec:multi_aspects} for the $i$-th document:
\begin{equation}
\hat{Y}_{i}=(\,\hat{y}_{i,1}\,..\,\hat{y}_{i,k}  \,), \quad Y_{i}^{algo}=(\,y_{i,1}^{algo}\,..\,y_{i,k}^{algo} \,)
\end{equation}
Each volume \texttt{V} is then calculated using the convex-hull algorithm and their overlap ($\sqcap$) is calculated using a shapely package\footnote{\url{https://pypi.org/project/Shapely/}}\footnote{Due to the lack of overlap calculation between two polygons of high dimensions, we reduce it to 2D PCA space.}.
The final \texttt{VO} is then:
\begin{equation}
\textsc{VO}_{algo} =\sum_{i=1}^{N}  \frac{\texttt{V}(\, E(Y_{i}^{algo})) \sqcap \texttt{V}\, ( E(\hat{Y}_{i}))}{\texttt{V}\,(E(\hat{Y}_{i}))} 
\end{equation}
where $N$ is the total number of input documents, 
$E$ is the BERT sentence encoder in Eq \ref{eq:bert}, and
$E(\hat{Y}_i)$ and $E(Y_{i}^{algo})$ are a set of vector representations of the reference and model summary sentences, respectively.
The volume overlap indicates how two summaries are semantically overlapped in a continuous embedding space.

\textbf{\texttt{Sentence Overlap (SO)} ratio.}
Even though ROUGE provides a recall-oriented lexical overlap, we don't know the upper-bound on  performance (called \texttt{oracle}) of the extractive summarization.
We extract the oracle extractive sentences (i.e.\ a set of input sentences) which maximizes ROUGE-L F-measure score with the reference summary. 
We then measure sentence overlap (\texttt{SO}) which determines how many extractive sentences from our algorithms are in the oracle summary.
The \texttt{SO} is: 
\begin{equation}
\textsc{SO}_{algo} =\sum_{i=1}^{n}  \frac{\texttt{C}( Y_{i}^{algo} \cap \hat{Y}_{i})}{\texttt{C}(\hat{Y}_{i})}
\end{equation}
where \texttt{C} is a function for counting the number of elements in a set.
The sentence overlap indicates how well the algorithm finds the oracle summaries for extractive summarization.

\begin{table*}[h]
\centering
\small
\begin{tabularx}{\textwidth}{@{}r@{\hskip 2mm}
c@{\hskip 1mm} c @{\hskip 2mm} 
c @{\hskip 1mm} c @{\hskip 2mm} 
c @{\hskip 1mm} c @{\hskip 2mm} 
c @{\hskip 1mm} c @{\hskip 2mm} 
c @{\hskip 1mm} c @{\hskip 2mm} 
c @{\hskip 1mm} c @{\hskip 2mm} 
c @{\hskip 1mm} c @{}}
\toprule
& \multicolumn{1}{c}{\texttt{CNNDM}} &  \multicolumn{1}{c}{\texttt{Newsroom}} &  \multicolumn{1}{c}{\texttt{Xsum}}&
\multicolumn{1}{c}{\texttt{PeerRead}} &
\multicolumn{1}{c}{\texttt{PubMed}} &
\multicolumn{1}{c}{\texttt{Reddit}} &
\multicolumn{1}{c}{\texttt{AMI}} &
\multicolumn{1}{c}{\texttt{BookSum}} & \multicolumn{1}{c}{\texttt{MScript}} 
 \\
\midrule
Source & \multicolumn{1}{c}{News} 
& \multicolumn{1}{c}{News} &  \multicolumn{1}{c}{News} &\multicolumn{1}{c}{Papers} &\multicolumn{1}{c}{Papers}
&\multicolumn{1}{c}{Post}&\multicolumn{1}{c}{Minutes}
&\multicolumn{1}{c}{Books}&\multicolumn{1}{c}{Script}
\\
Multi-sents. & \multicolumn{1}{c}{\checkmark} 
& \multicolumn{1}{c}{\checkmark}  
&\multicolumn{1}{c}{X} 
&\multicolumn{1}{c}{\checkmark} &\multicolumn{1}{c}{\checkmark} 
&\multicolumn{1}{c}{X} 
&\multicolumn{1}{c}{\checkmark}
&\multicolumn{1}{c}{\checkmark}
& \multicolumn{1}{c}{\checkmark}  
\\
\midrule
Data size &287K/11K	&992K/109K	&203K/11K &10K/550 &	 21K/2.5K & 404/48 &  98/20 & - /53 &- /1K \\
\midrule
Avg src sents. &40/34& 24/24&33/33&45/45&97/97 &19/15 & 767/761	&  - /6.7K& - /3K& \\
Avg tgt sents. &4/4	& 1.4/1.4 &	1/1	& 6/6 &10/10 & 1/1 & 17/17& - /336& - /5 \\
\midrule
Avg src tokens & 792/779 &	769	/762 &	440/442 &	1K/1K  &2.4K/2.3K & 296/236 &6.1K/6.4K &- /117K & - /23.4K \\
Avg tgt tokens & 55/58 & 30/31&	23/23 &	144/146&258/258 & 24/25 &281/277& - /6.6K & - /104 \\
\bottomrule
\end{tabularx}
\caption{\label{tab:dataset} Data statistics on summarization corpora. 
Source is the domain of dataset.
Multi-sents. is whether the summaries are multiple sentences or not.
All statistics are divided by Train/Test except for \texttt{BookSum} and \texttt{MScript}.
}
\end{table*}

\section{Summarization Corpora}\label{sec:data}

We use various domains of summarization datasets to conduct the bias analysis across corpora and systems.
Each dataset has source documents and corresponding abstractive target summaries. 
We provide a list of datasets used along with a brief description and our pre-processing scheme:

\begin{itemize}[noitemsep,topsep=0pt,leftmargin=*]
\item \texttt{CNNDM} \cite{nallapatiabstractive}: contains 300K number of online news articles. It has multiple sentences (4.0 on average) as a summary.
\item \texttt{Newsroom} \cite{N18-1065}: contains 1.3M news articles and written summaries by authors and editors from 1998 to 2017. It has both extractive and abstractive summaries.
\item \texttt{XSum} \cite{narayan2018don}: has news articles and their single but abstractive sentence summaries mostly written by the original author.
\item \texttt{PeerRead} \cite{kang18naacl}: consists of scientific paper drafts in top-tier computer science venues as well as \url{arxiv.org}. We use full text of introduction section as source document and of abstract section as target summaries.
\item \texttt{PubMed} \cite{kedzie2018content}: is 25,000 medical journal papers from the PubMed Open Access Subset.\footnote{\url{https://www.ncbi.nlm.nih.gov/pmc/tools/openftlist/}} 
Unlike \texttt{PeerRead}, full paper except for abstract is used as source documents. 
\item \texttt{MScript} \cite{gorinski2015movie}: is a collection of movie scripts from ScriptBase corpus and their corresponding user summaries of the movies.
\item \texttt{BookSum} \cite{mihalcea2007explorations}: is a dataset of classic books paired to summaries from Grade Saver\footnote{\url{http://www.gradesaver.com}} and Cliff’s Notes\footnote{\url{http://www.cliffsnotes.com/}}. 
Due to a large number of sentences, we only choose the first 1K sentences for source document and the first 50 sentences for target summaries. 
\item \texttt{Reddit} \cite{ouyang2017crowd}: is a collection of personal posts from \url{reddit.com}. We use a single abstractive summary per post. The same data split from \citet{kedzie2018content} is used.
\item \texttt{AMI} \cite{carletta2005ami}: is documented meeting minutes from a hundred hours of recordings and their abstractive summaries. 
\end{itemize}

Table \ref{tab:dataset} summarizes the characteristics of each dataset.
We note that the Gigaword \cite{graff2003english}, New York Times\footnote{\url{https://catalog.ldc.upenn.edu/LDC2008T19}}, and Document Understanding Conference (DUC)\footnote{\url{http://duc.nist.gov}} are also popular datasets commonly used in summarization analyses, though here we exclude them as they represent only additional collections of news articles, showing similar tendencies to the other news datasets such as \texttt{CNNDM}.


\section{Analysis on  Corpus Bias}\label{sec:analysis_corpora}
We conduct different analyses of how each corpus is biased with respect to the sub-aspects. We highlight some key findings for each sub-section.

\begin{table*}[h!]
\fontsize{7.6}{9.0}\selectfont
\centering
\vspace{0mm}

\begin{tabularx}{\textwidth}{
@{}c@{}|@{\hskip 2mm}r@{}
@{\hskip 2mm} c@{\hskip 1mm}c@{\hskip 1mm}c  
@{\hskip 2mm} c@{\hskip 1mm}c@{\hskip 1mm}c 
@{\hskip 2mm} c@{}c@{}c 
@{\hskip 2mm} c@{\hskip 1mm}c@{\hskip 1mm}c
@{\hskip 2mm} c@{\hskip 1mm}c@{\hskip 1mm}c  
@{\hskip 2mm} c@{}c@{}c 
@{\hskip 2mm} c@{\hskip 1mm}c@{\hskip 1mm}c
@{\hskip 2mm} c@{\hskip 1mm}c@{\hskip 1mm}c
@{\hskip 2mm} c@{\hskip 1mm}c@{\hskip 1mm}c
@{}}
\toprule
&& \multicolumn{3}{c}{\texttt{CNNDM}} &  \multicolumn{3}{c}{\texttt{NewsRoom}} & \multicolumn{3}{c}{\texttt{XSum}} &
\multicolumn{3}{c}{\texttt{PeerRead}} & \multicolumn{3}{c}{\texttt{PubMed}} 
&\multicolumn{3}{c}{\texttt{Reddit}} 
&\multicolumn{3}{c}{\texttt{AMI}}
&\multicolumn{3}{c}{\texttt{BookSum}} 
&\multicolumn{3}{c}{\texttt{MScript}}\\
\cmidrule(lr){3-5} \cmidrule(lr){6-8} \cmidrule(lr){9-11} \cmidrule(lr){12-14} \cmidrule(lr){15-17} \cmidrule(lr){18-20} \cmidrule(lr){21-23} \cmidrule(lr){24-26}  \cmidrule(lr){27-29}
&& \textsc{R} & VO & SO 
& \textsc{R} & VO & SO 
& \textsc{R} & VO & SO 
& \textsc{R} & VO & SO 
& \textsc{R} & VO & SO 
& \textsc{R} & VO & SO 
& \textsc{R} & VO & SO 
& \textsc{R} & VO & SO 
& \textsc{R} & VO & SO  \\
\midrule
&\textsc{Random}&19.1&18.6&14.6&10.1&2.1&9.0&9.3&-&8.4&27.9&42.5&26.2&30.1&46.9&13.0&11.8&-&11.3&12.0&39.3&2.4&29.4&85.8&4.9&8.1&25.2&0.1\\
&\textsc{Oracle} &42.8&-&-&48.1&-&-&19.6&-&-&46.3&-&-  &47.0&-&-&30.0&-&-&32.0&-&-&38.9&-&-&24.2&-&-\\
\midrule
\parbox[t]{1mm}{\multirow{3}{*}{\rotatebox[origin=c]{90}{{\tiny{\textsc{Position}}}}}} &
First&\textbf{30.7}&13.1&\textbf{30.7}&\textbf{32.2}&\textbf{4.4}&\textbf{37.8}&9.1&-&8.7&\textbf{32.0}&40.7&\textbf{30.3} &27.6&44.3&13.8&\textbf{15.3}&-&\textbf{19.9} & 11.4&48.0&\textbf{3.8}&\textbf{29.1}&85.1&\textbf{7.4}&6.9&12.4&\textbf{0.7}\\
&Last&16.4&18.6&8.2&7.7&1.9&4.4&8.3&-&7.0&28.9&38.5&27.0 &28.9&45.2&14.0&11.2&-&10.7&7.8&42.1&2.0&26.5&85.3&3.3&\textbf{8.8}&19.5&0.2\\
&Middle&21.5&18.7&11.8&12.4&1.9&5.6&9.1&-&9.1&29.7&40.7&22.8 &28.9&45.9&12.3&11.5&-&7.1&11.1&36.4&2.3&27.9&83.0&4.9&8.0&23.9&0.1\\
\midrule
\parbox[t]{1mm}{\multirow{2}{*}{\rotatebox[origin=c]{90}{{\tiny{\textsc{Divers.}}}}}} &
ConvFall&21.6&\textbf{57.7}&15.0&10.6&4.2&7.3&8.4&-&8.0&29.8&\textbf{77.5}&25.9 &28.2&\textbf{93.5}&11.2&11.6&-&7.5&\textbf{14.0}&\textbf{98.6}&2.4&16.9&\textbf{99.7}&2.2&8.5&\textbf{59.2}&0.2\\
&Heuris.&21.4&19.8&14.6&10.5&2.4&7.6&8.4&-&8.1&29.2&36.6&24.8 &27.5&59.7&10.5&11.5&-&7.1&10.7&66.0&2.4&26.9&99.7&4.5&6.4&5.7&0.2\\
\midrule
\parbox[t]{1mm}{\multirow{2}{*}{\rotatebox[origin=c]{90}{{\tiny{\textsc{Import.}}}}}} &
NNear.&22.0&3.3&16.6&13.5&0.5&10.0&\textbf{9.8}&-&\textbf{10.1}&30.6&8.4&26.7 &\textbf{31.8}&9.3&\textbf{15.5}&13.8&-&12.2&1.3&0.2&0.1&27.9&1.5&5.1&8.7&0.9&0.3\\
&KNear.&23.0&3.9&17.7&14.0&0.7&10.9&9.3&-&9.1&30.6&9.9&27.0 &29.6&10.5&15.0&10.4&-&8.5&0.0&0.1&0.0&21.8&1.4&3.7&0.6&0.0&0.1\\
\bottomrule
\end{tabularx}
\caption{\label{tab:analysis_multi} Comparison of different corpora w.r.t the three sub-aspects: \textsc{position}, \textsc{diversity}, and \textsc{importance}.
We averaged R1, R2, and RL as \textsc{R} (See Appendix for full scores). 
 Note that volume overlap (\texttt{VO}) doesn't exist when target summary has a single sentence. (i.e., \texttt{XSum}, \texttt{Reddit})
 }
\end{table*}

\subsection{Multi-aspect analysis}\label{sec:multi_analysis_corpora}
Table \ref{tab:analysis_multi} shows a comparison of the three aspects for each corpus where we include random selection and the oracle set.
For each dataset metrics are calculated on a test set except for \texttt{BookSum} and \texttt{AMI} where we use train+test due to the smaller sample size. 

\textbf{Earlier isn't always better.} Sentences selected early in the source show high \texttt{ROUGE} and \texttt{SO} on \texttt{CNNDM}, \texttt{Newsroom}, \texttt{Reddit}, and \texttt{BookSum}, but not in other domains such as medial journals and meeting minutes, and the condensed news summaries (\texttt{XSum}). 
For summarization of movie scripts in particular, the last sentences seem to provide more important summaries.

\textbf{\texttt{XSum} requires much \textsc{importance} than other corpora.} Interestingly, the most powerful algorithm for \texttt{XSum} is \texttt{N-Nearest}. This shows that summaries in \texttt{XSum} are indeed collected by abstracting multiple important contents into single sentence, avoiding the position bias.

\textbf{First, ConvexFall, and N-Nearest tend to work better than the other algorithms for each aspect.} \texttt{First} is better than \texttt{Last} or \texttt{Middle} in new articles except for \texttt{XSum} and personal posts, while not in academic papers (i.e., \texttt{PeerRead}, \texttt{PubMed}) and meeting minutes.
\texttt{ConvexFall} finds the set of sentences that maximize the semantic volume overlap with the target sentences better than the heuristic one.

\textbf{\texttt{ROUGE} and \texttt{SO} show similar behavior, while \texttt{VO} does not.} In most evaluations, ROUGE scores are linear to SO ratios as expected. However, VO has high variance across algorithms and aspects. This is mainly because the semantic volume assumption maximizes the semantic diversity, but sacrifices other aspects like importance by choosing the outlier sentences over the convex hull.

\textbf{Social posts and news articles are biased to the position aspect while the other two aspects appear less relevant.} (Figure \ref{fig:triangle_all} (a)) However, \texttt{XSum} requires all aspects equally but with relatively less relevant to any of aspects than the other news corpora. 

\textbf{Paper summarization is a well-balanced task.} The variance of \texttt{SO} across the three aspects in \texttt{PeerRead} and \texttt{PubMed} is relatively smaller than other corpora.
This indicates that abstract summary of the input paper requires the three aspects at the same time.
\texttt{PeerRead} has relatively higher \texttt{SO} then \texttt{PubMed} because it only summarize text in Introduction section, while \texttt{PubMed} summarize whole paper text, which is much difficult (almost random performance).

\textbf{Conversation, movie script and book summarization are very challenging.}
Conversation of spoken meeting minutes includes a lot of witty replies repeatedly (e.g., `okay.' , `mm -hmm.' , `yeah.'), causing importance and diversity measures to suffer.
\texttt{MScript} and \texttt{BookSum} which include very long input document seem to be extremely difficult task, showing almost random performance.

\subsection{Intersection between the sub-aspects}

\begin{figure}[h!]
\centering
{
\subfloat[{\texttt{CNNDM} (49.4\%)}]{
\includegraphics[trim=0cm 0cm 0cm 1cm,clip,valign=b,width=.49\linewidth]{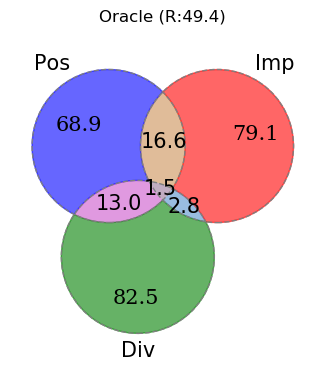}}
\subfloat[{\texttt{XSum} (76.8\%)}]{
\includegraphics[trim=0cm 0cm 0cm 1cm,clip,valign=b,width=.49\linewidth]{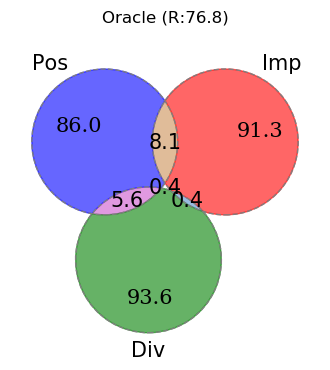}}
\vspace{-3mm}
\\
\subfloat[{\texttt{PeerRead} (37.6\%)}]{
\includegraphics[trim=0cm 0cm 0cm 1cm,clip,valign=b,width=.49\linewidth]{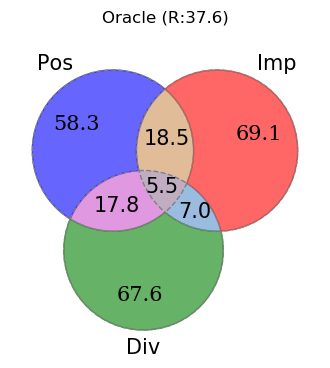}}
\subfloat[{\texttt{Reddit} (68.1\%)}]{
\includegraphics[trim=0cm 0cm 0cm 1cm,clip,valign=b,width=.49\linewidth]{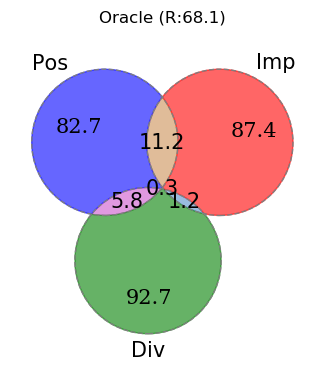}}
\vspace{-3mm}\\
\subfloat[{\texttt{AMI} (94.1\%)}]{
\includegraphics[trim=0cm 0cm 0cm 1cm,clip,valign=b,width=.49\linewidth]{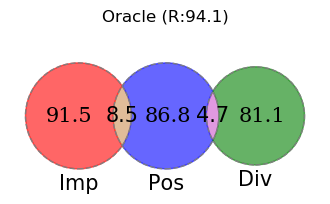}}
\subfloat[{\texttt{BookSum} (87.1\%)}]{
\includegraphics[trim=0cm 0cm 0cm 1cm,clip,valign=b,width=.49\linewidth]{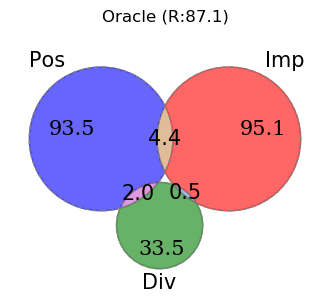}}
}
\caption{\label{fig:venndiagram} Intersection of averaged summary sentence overlaps across the sub-aspects.
We use \texttt{First} for \textsc{Position}, \texttt{ConvexFall} for \textsc{Diversity}, and \texttt{N-Nearest} for \textsc{Importance}.
The number in the parenthesis called \textit{Oracle Recall} is the averaged ratio of how many the oracle sentences are \textbf{NOT} chosen by union set of the three sub-aspect algorithms.
Other corpora are in Appendix with their Oracle Recalls: \texttt{Newsroom}(54.4\%), \texttt{PubMed} (64.0\%) and \texttt{MScript}  (99.1\%).
\vspace{-3mm}
}
\end{figure}

\begin{figure}[ht]
\vspace{0mm}
\centering
{

{
\includegraphics[trim=0.8cm 1.5cm 0.8cm 2.4cm,clip,width=.88\linewidth]{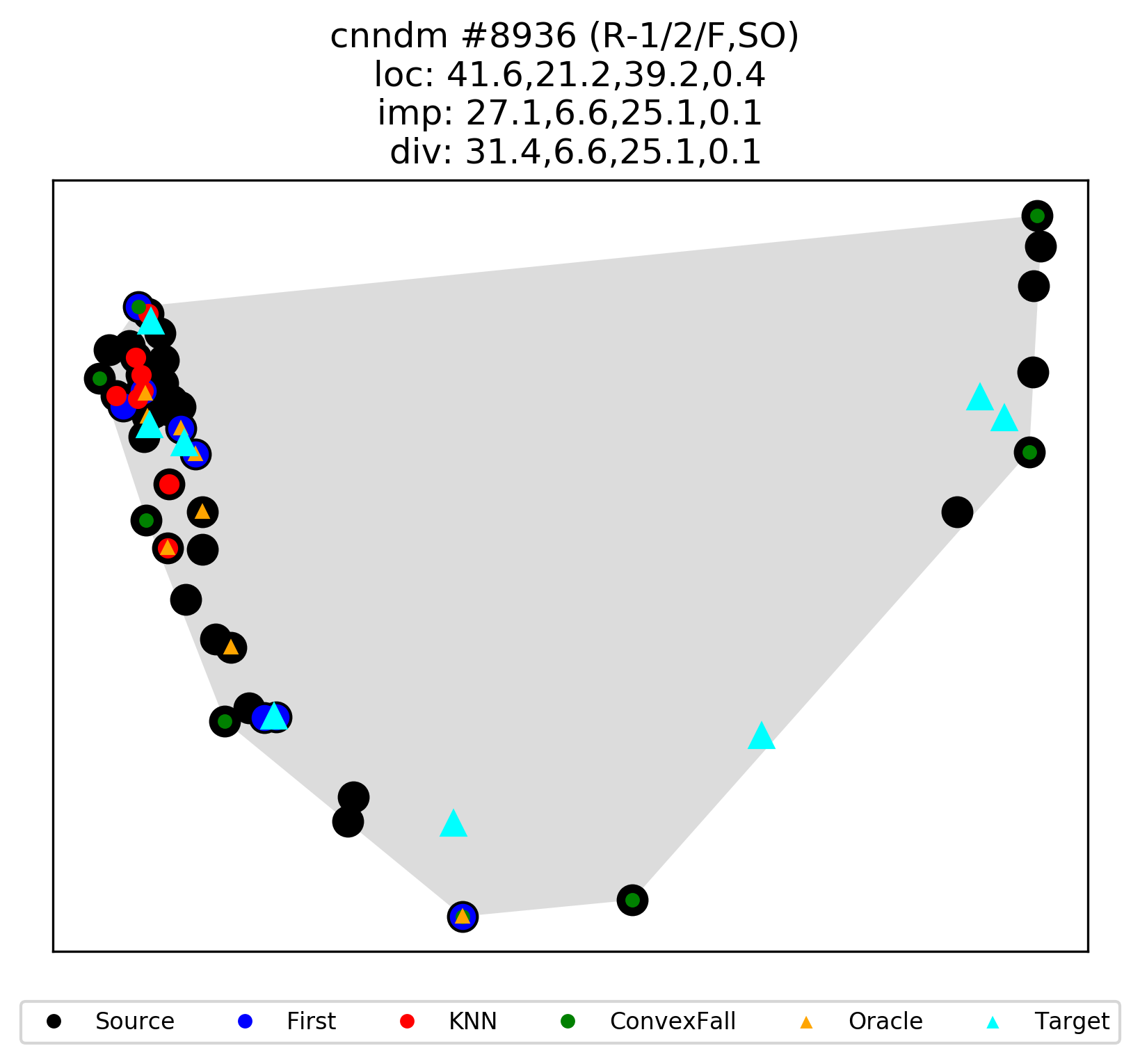}
}
\\
}
\caption{\label{fig:pca_multi_aspect} PCA projection of extractive summaries chosen by multiple aspects of algorithms (\texttt{CNNDM}). Source and target sentences are black circles ($\newmoon$) and {\color{cyan} cyan} triangles, respectively. The {\color{blue}blue}, {\color{green}green}, {\color{red}red} circles are summary sentences chosen by \texttt{First}, \texttt{ConvexFall}, \texttt{NN}, respectively.
The {\color{yellow}yellow} triangles are the oracle sentences. 
Shaded polygon represents a ConvexHull volume of sample source document.
Best viewed in color.
Please find more examples in Appendix.
\vspace{-3mm}
}
\end{figure}

\begin{figure*}[h]
\vspace{0mm}
\centering
\footnotesize
\begin{tabularx}{\textwidth}{YYYYYYYYY}
\texttt{CNNDM} & \texttt{NewsRoom} & \texttt{XSum} & \texttt{PeerRead} &  \texttt{PubMed}  & \texttt{Reddit} & \texttt{AMI} & \texttt{BookSum} & \texttt{MScript}\\
\end{tabularx}
\newline
\vspace*{-0.7cm}
\newline
{
\subfloat[\footnotesize{\textsc{{Position}}}]{
\includegraphics[trim=0.9cm 0.9cm 1.5cm 1.3cm,clip,width=.105\linewidth]{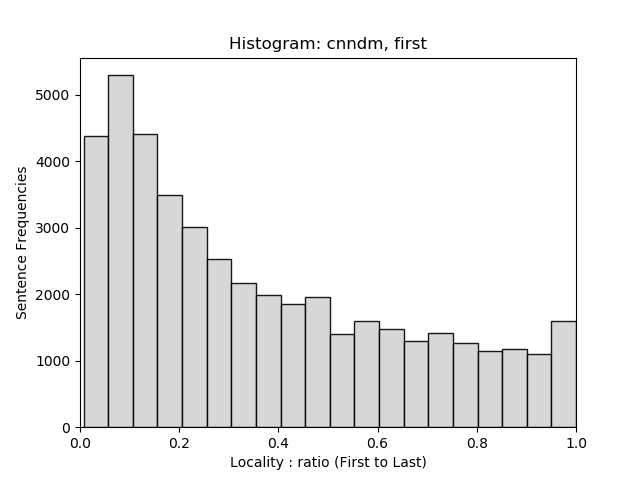}
\includegraphics[trim=0.9cm 0.9cm 1.5cm 1.3cm,clip,width=.105\linewidth]{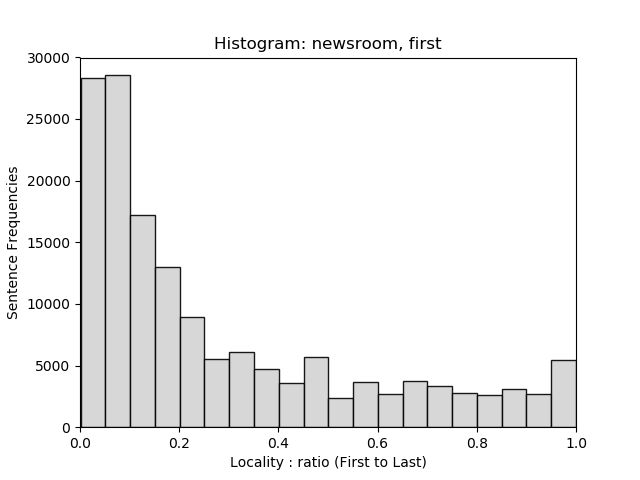}
\includegraphics[trim=0.9cm 0.9cm 1.5cm 1.3cm,clip,width=.105\linewidth]{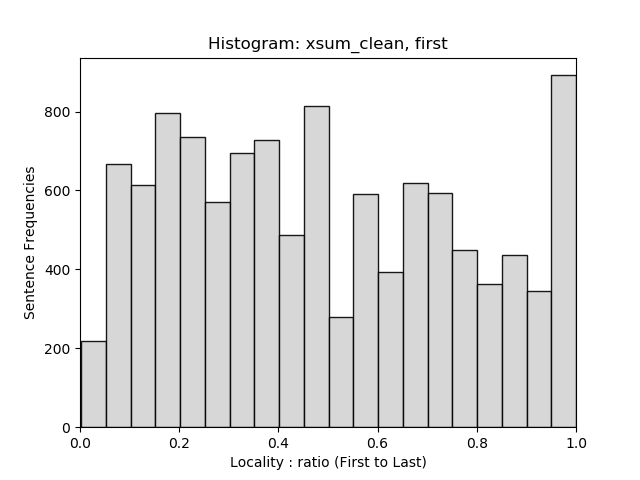}
\includegraphics[trim=0.9cm 0.9cm 1.5cm 1.3cm,clip,width=.105\linewidth]{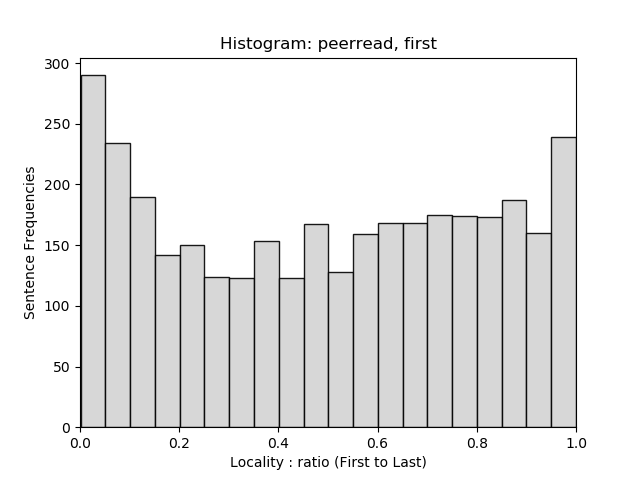}
\includegraphics[trim=0.9cm 0.9cm 1.5cm 1.3cm,clip,width=.105\linewidth]{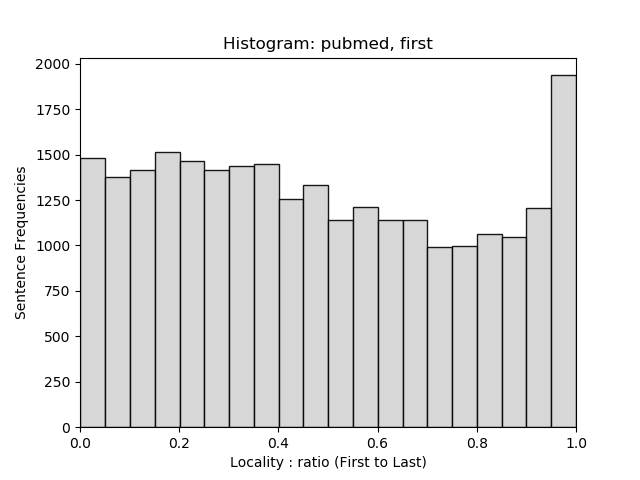}
\includegraphics[trim=0.9cm 0.9cm 1.5cm 1.3cm,clip,width=.105\linewidth]{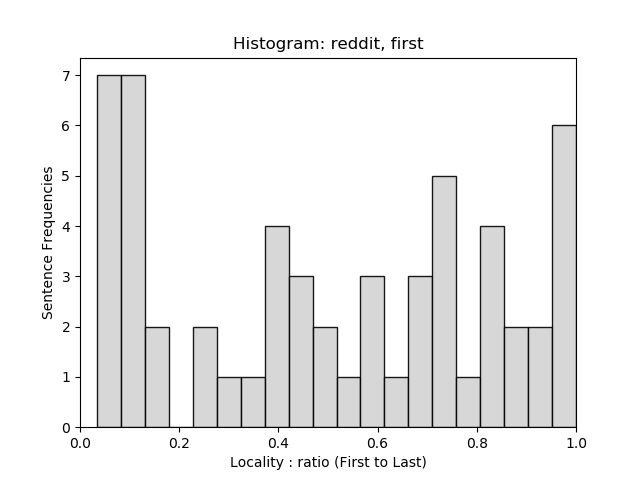}
\includegraphics[trim=0.9cm 0.9cm 1.5cm 1.3cm,clip,width=.105\linewidth]{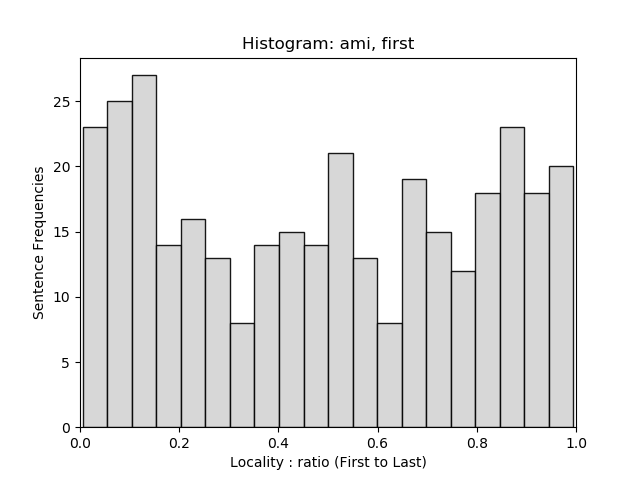}
\includegraphics[trim=0.9cm 0.9cm 1.5cm 1.3cm,clip,width=.105\linewidth]{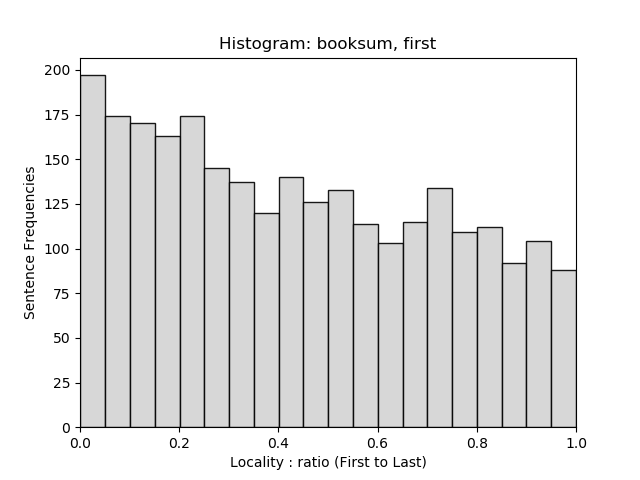}
\includegraphics[trim=0.9cm 0.9cm 1.5cm 1.3cm,clip,width=.105\linewidth]{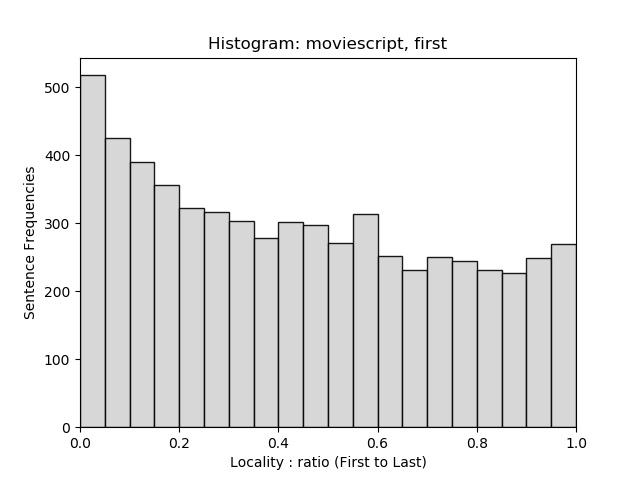}
}\vspace{-2mm}
\\
\subfloat[\footnotesize{\textsc{{Diversity}}}]{
\includegraphics[trim=0.9cm 0.9cm 1.5cm 1.3cm,clip,width=.105\linewidth]{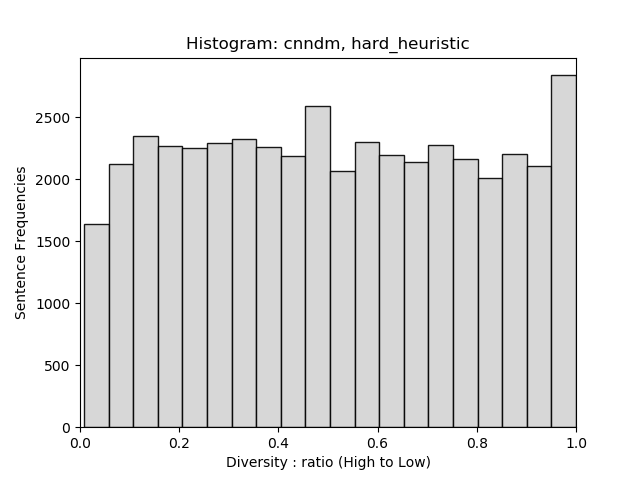}
\includegraphics[trim=0.9cm 0.9cm 1.5cm 1.3cm,clip,width=.105\linewidth]{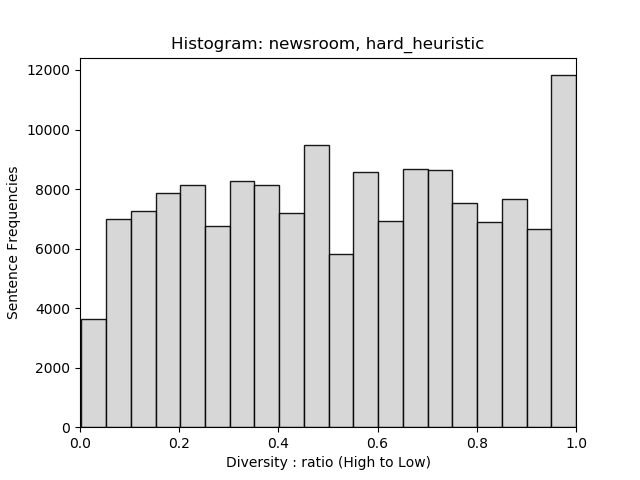}
\includegraphics[trim=0.9cm 0.9cm 1.5cm 1.3cm,clip,width=.105\linewidth]{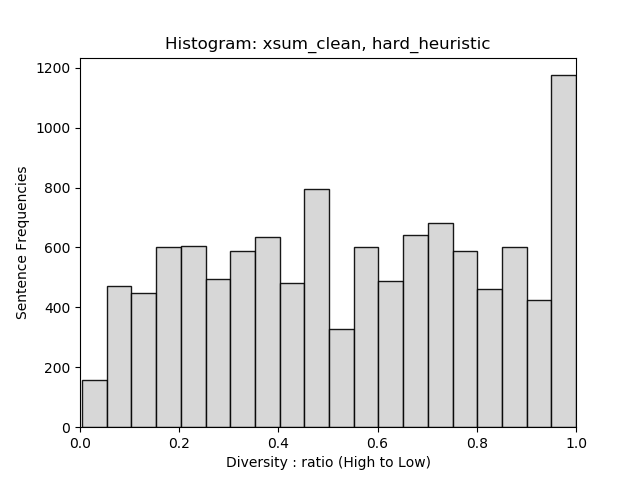}
\includegraphics[trim=0.9cm 0.9cm 1.5cm 1.3cm,clip,width=.105\linewidth]{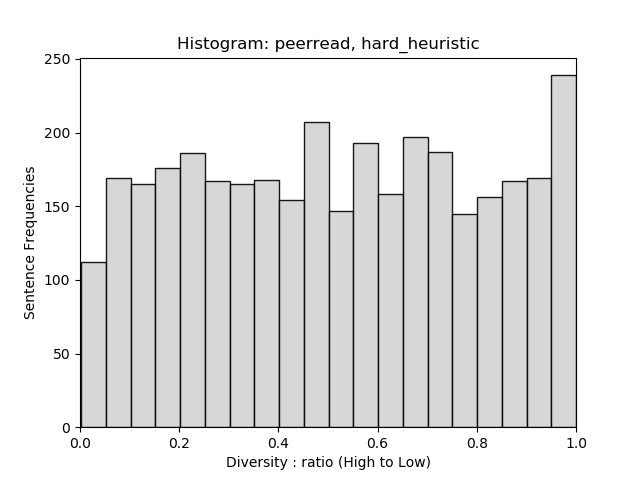}
\includegraphics[trim=0.9cm 0.9cm 1.5cm 1.3cm,clip,width=.105\linewidth]{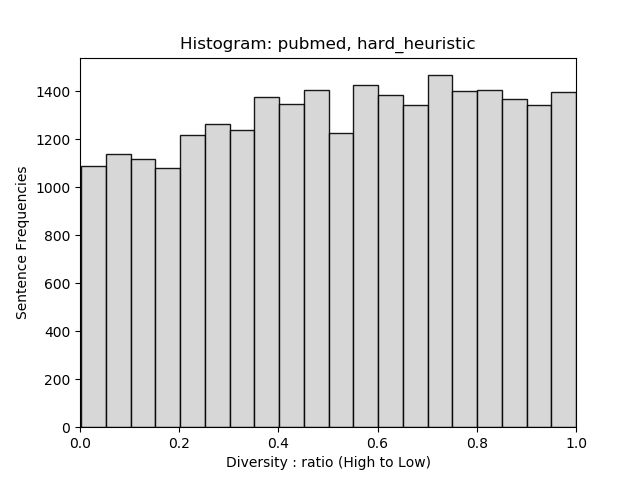}
\includegraphics[trim=0.9cm 0.9cm 1.5cm 1.3cm,clip,width=.105\linewidth]{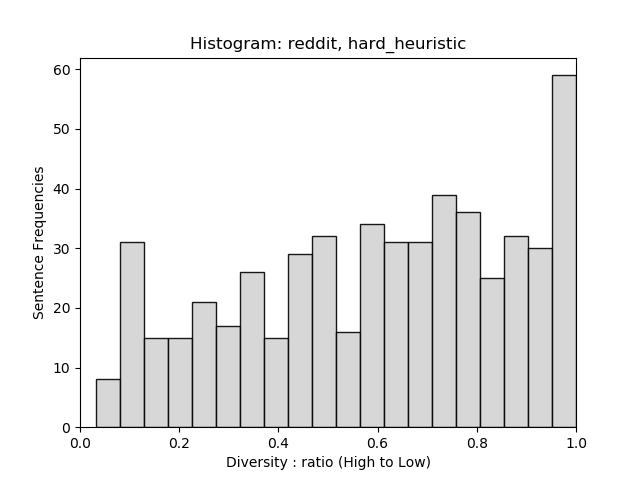}
\includegraphics[trim=0.9cm 0.9cm 1.5cm 1.3cm,clip,width=.105\linewidth]{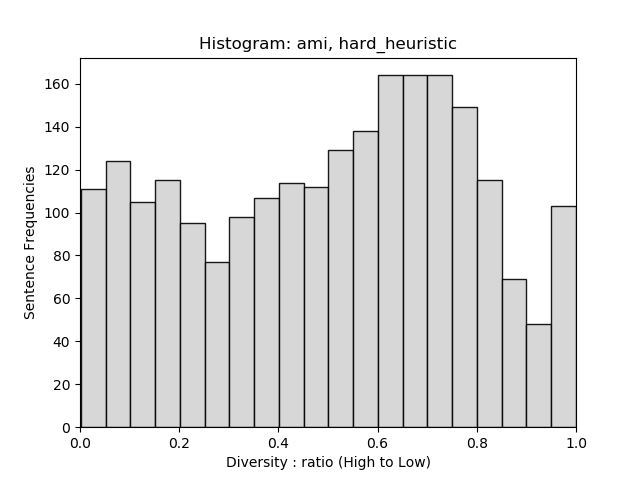}
\includegraphics[trim=0.9cm 0.9cm 1.5cm 1.3cm,clip,width=.105\linewidth]{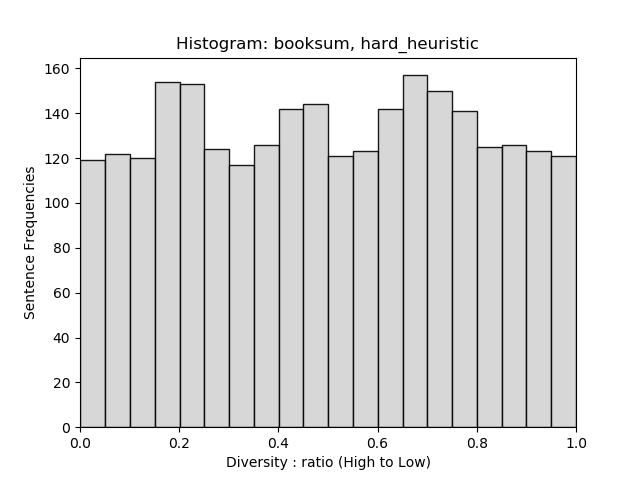}
\includegraphics[trim=0.9cm 0.9cm 1.5cm 1.3cm,clip,width=.105\linewidth]{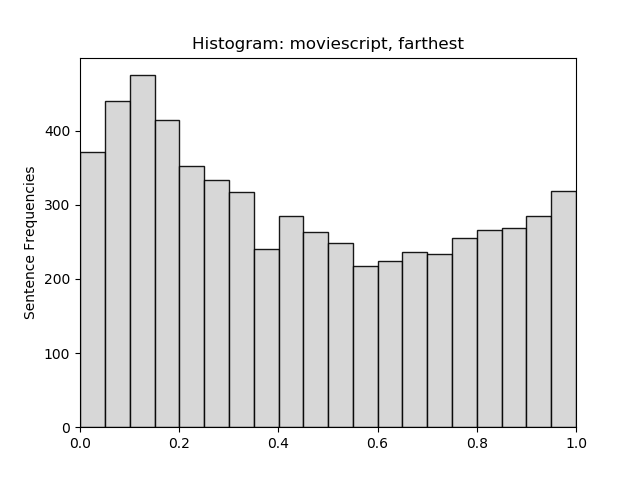}
}\vspace{-2mm}
\\
 \subfloat[\footnotesize{\textsc{{Importance}}}]{
 \includegraphics[trim=0.9cm 0.9cm 1.5cm 1.3cm,clip,width=.105\linewidth]{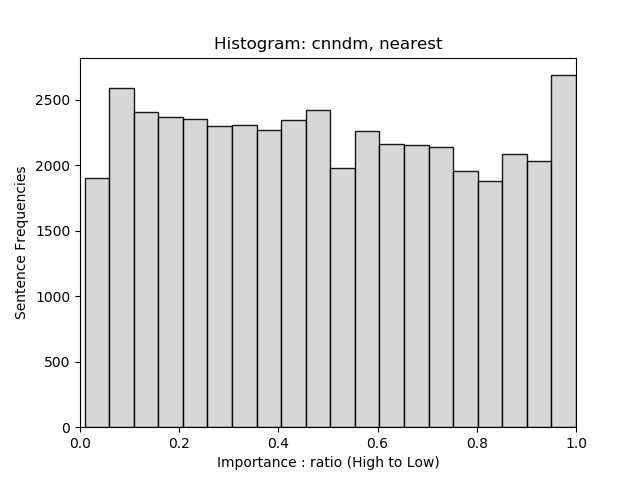}
 \includegraphics[trim=0.9cm 0.9cm 1.5cm 1.3cm,clip,width=.105\linewidth]{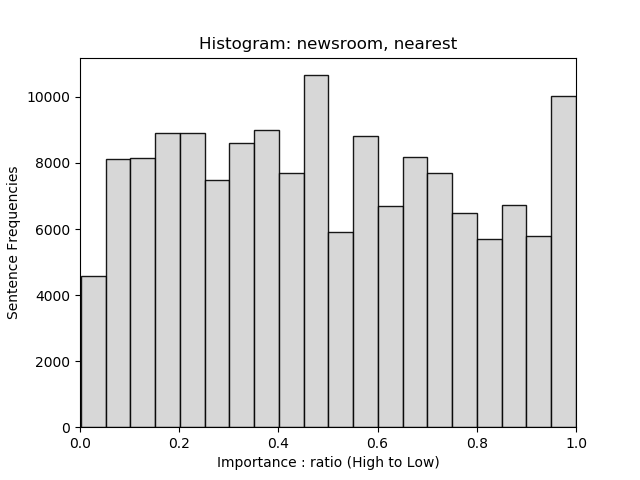}
 \includegraphics[trim=0.9cm 0.9cm 1.5cm 1.3cm,clip,width=.105\linewidth]{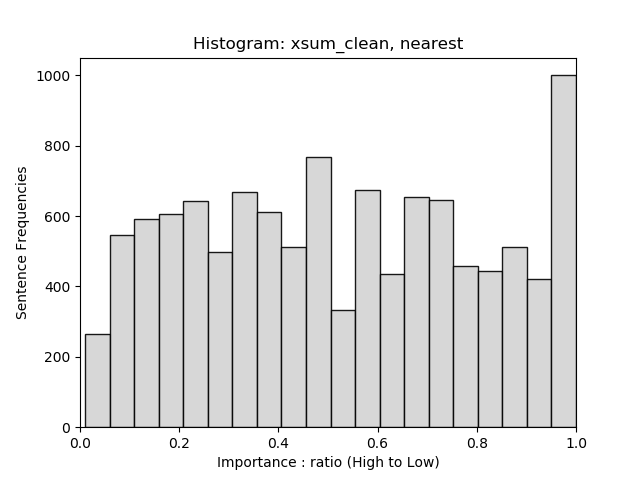}
 \includegraphics[trim=0.9cm 0.9cm 1.5cm 1.3cm,clip,width=.105\linewidth]{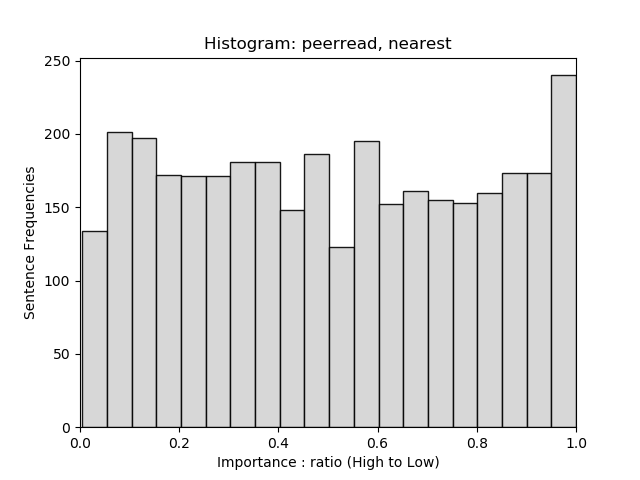}
 \includegraphics[trim=0.9cm 0.9cm 1.5cm 1.3cm,clip,width=.105\linewidth]{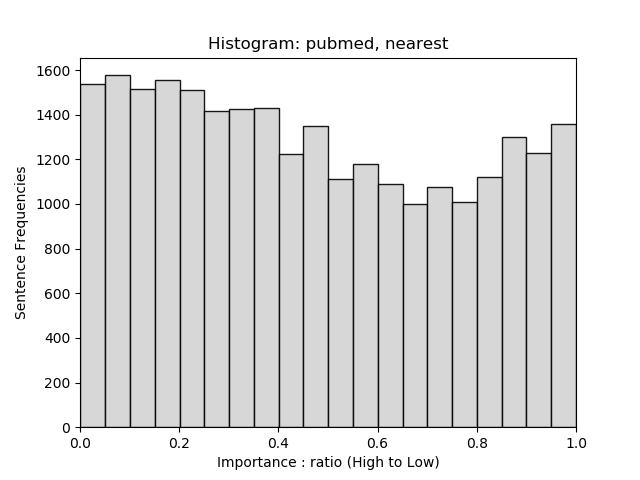}
 \includegraphics[trim=0.9cm 0.9cm 1.5cm 1.3cm,clip,width=.105\linewidth]{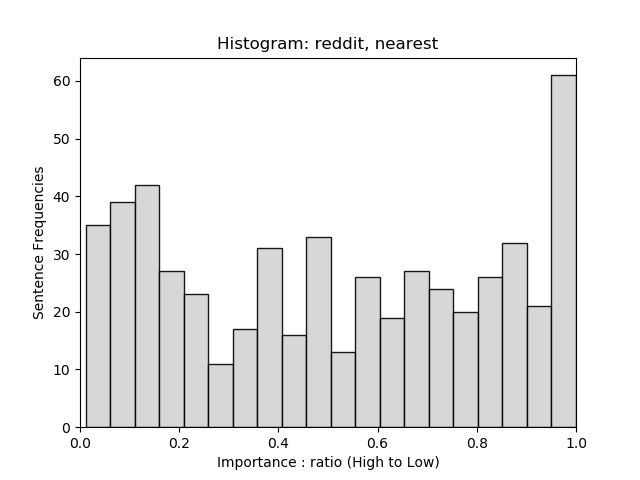}
 \includegraphics[trim=0.9cm 0.9cm 1.5cm 1.3cm,clip,width=.105\linewidth]{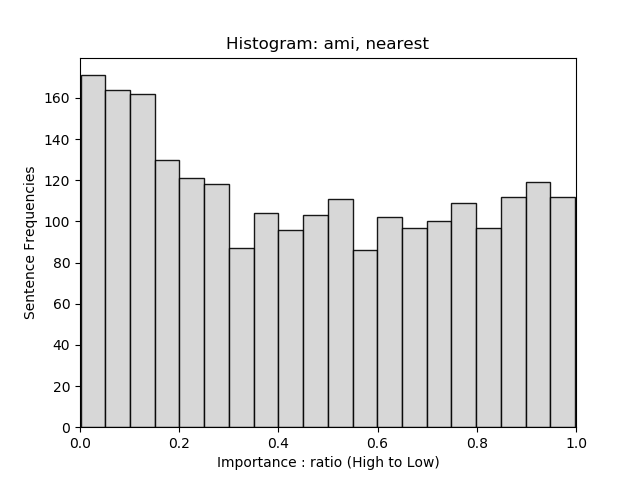}
 \includegraphics[trim=0.9cm 0.9cm 1.5cm 1.3cm,clip,width=.105\linewidth]{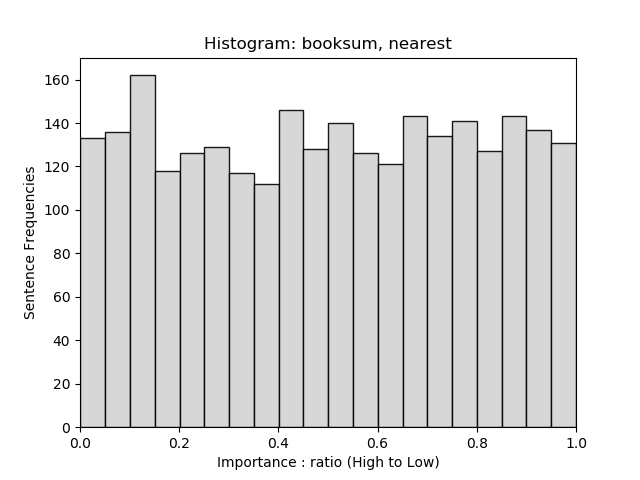}
  \includegraphics[trim=0.9cm 0.9cm 1.5cm 1.3cm,clip,width=.105\linewidth]{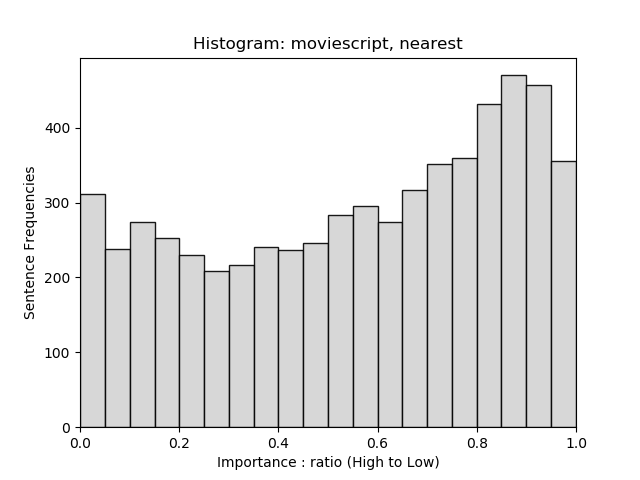}}
}
\caption{\label{fig:hist1d} Sentence overlap proportion of each sub-aspect (row) with the oracle summary across corpora (column).
y-axis is the frequency of overlapped sentences with the oracle summary.
X-axis is the normalized RANK of individual sentences in the input document where size of bin is 0.05. 
E.g., the first / the most diverse / the most important sentence is in the first bin.
If earlier bars are frequent, the aspect is positively relevant to the corpus.
}
\vspace{0mm}
\end{figure*}

Averaged ratios across the sub-aspects do not capture how the actual summaries overlap with each other.
Figure \ref{fig:venndiagram} shows Venn diagrams of how sets of summary sentences chosen by different sub-aspects are overlapped each other on average.

\textbf{\texttt{XSum}, \texttt{BookSum}, and \texttt{AMI} have high Oracle Recall.} If we develop a mixture model of the three aspects, the Oracle Recall means its upper bound, meaning that another sub-aspect should be considered regardless of the mixture model.
This indicates that existing procedures are not enough to cover the Oracle sentences.
For example, \texttt{AMI} and \texttt{BookSum} have a lot of repeated noisy sentences, some of which could likely be removed without a significant loss of pertinent information.

\textbf{\textsc{Importance} and \textsc{Diversity} are less overlapped with each other.} This means that important sentences are not always diverse sentences, indicating that they should be considered together.

\subsection{Summaries in a embedding space}

Figure \ref{fig:pca_multi_aspect} shows two dimensional PCA projections of a document in \texttt{CNNDM} on the embedding space.  

\textbf{Source sentences are clustered on the convexhull border, not in the middle.} 
We conjecture that sentences are not uniformly distributed in the embedding space but their positions gradually move over the convexhull.
Target summaries reflect different sub-aspects according to the sample and corpora. For example, many target sentences in \texttt{CNNDM} are near by \texttt{First-k} sentences. 
\subsection{Single-aspect analysis}\label{sec:single_analysis_corpora}


We calculate the frequency of source sentences overlapped with the oracle summary where the source sentences are ranked differently according to the algorithm of each aspect (See Figure \ref{fig:hist1d}). 
Heavily skewed histograms indicate that oracle sentences are positively (right-skewed) or negatively (left-skewed) related to the sub-aspect.

In most cases, some oracle sentences are overlapped to the first part of the source sentences.
Even though their degrees are different, oracle summaries from many corpora (i.e, \texttt{CNNDM}, \texttt{NewsRoom}, \texttt{PeerRead}, \texttt{BookSum}, \texttt{MScript}) are highly related to the \textsc{position}.
Compared to the other corpora, \texttt{PubMed} and \texttt{AMI} contain more top-ranked important sentences in their oracle summaries.
News articles and papers tend to find oracle sentences without \textsc{diversity} (i.e., right-skewed), meaning that non-diverse sentences are frequently selected as part of the oracle.

We also measure how many \textit{new} words occur in abstractive target summaries, by comparing overlap between oracle summaries and document sentences (Table \ref{tab:ngram_oracle}).
One thing to note is that \texttt{XSum} and \texttt{AMI} have less \textit{new} words in their target summaries.
On the other hand, paper datasets (i.e., \texttt{PeerRead} and \texttt{PubMed}) include a lot, indicating that abstract text in academic paper is indeed ``abstract''.




\begin{table}[ht]
\centering
\small
\vspace{0mm}
\begin{tabularx}{.99\linewidth}{
@{} rYYYYY @{}}
\toprule
& \multicolumn{1}{@{}c@{}}{\textbf{R(\textsc{O},\textsc{T})}} & 
\multicolumn{2}{@{}c@{}}{\textbf{\textsc{O}$\cap$\textsc{T}}} & \multicolumn{2}{@{}c@{}}{\textbf{\textsc{T}$\setminus$\textsc{S}}} \\
\cmidrule(lr){2-2} \cmidrule(lr){3-4} \cmidrule(lr){5-6}
& 
& Unigram & Bigram
& Unigram & Bigram  \\
\midrule
\texttt{CNNDM}&42.8&66.0&36.4&14.7&5.7\\
\texttt{Newsroom}&48.1&60.7&43.4&7.8&3.4\\
\texttt{XSum}&19.6&30.4&6.9&8.4&1.2\\
\texttt{PeerRead}&46.3&48.5&27.2&20.1&8.8\\
\texttt{PubMed}&47.0&52.1&27.7&16.7&6.7\\
\texttt{Reddit}&30.0&41.0&16.4&13.8&3.8\\
\texttt{AMI}&32.0&28.1&8.5&10.6&1.5\\
\texttt{BookSum}&38.9&25.6&8.9&6.7&1.7\\
\texttt{MScript}&38.9&13.9&4.0&0.3&0.1\\

\bottomrule
\end{tabularx}
\vspace{0mm}
\caption{\label{tab:ngram_oracle} 
ROUGE of oracle summaries and averaged N-gram overlap ratios. 
\textsc{O}, \textsc{T} and \textsc{S} are a set of N-grams from \textsc{Oracle}, \textsc{Target} and \textsc{Source} document, respectively.
\textbf{R(\textsc{O},\textsc{T})} is the averaged ROUGE between oracle and target summaries, showing how similar they are.
\textbf{\textsc{O}$\cap$\textsc{T}} shows N-gram overlap between oracle and target summaries. 
The higher the more overlapped words in between. 
\textbf{\texttt{\textsc{T}$\setminus$\textsc{S}}} is a proportion of N-grams in target summaries not occurred in source document. 
The lower the more abstractive (i.e., new words) target summaries. 
}
\vspace{-3mm}
\end{table}

\begin{table*}[htbp!]
\fontsize{6.5}{7.5}\selectfont
\centering
\vspace{0mm}

\begin{tabularx}{\linewidth}{
@{}c@{}|@{\hskip 1.5mm}r
@{\hskip 1.5mm} c@{\hskip 0.8mm}c@{\hskip 0.8mm}c 
@{\hskip 1.5mm} c@{\hskip 0.8mm}c@{\hskip 0.8mm}c
@{\hskip 1.5mm} c@{\hskip 0.8mm}c@{\hskip 0.8mm}c
@{\hskip 1.5mm} c@{\hskip 0.8mm}c@{\hskip 0.8mm}c
@{\hskip 1.5mm} c@{\hskip 0.8mm}c@{\hskip 0.8mm}c
@{\hskip 1.5mm} c@{\hskip 0.8mm}c@{\hskip 0.8mm}c
@{\hskip 1.5mm} c@{\hskip 0.8mm}c@{\hskip 0.8mm}c
@{\hskip 1.5mm} c@{\hskip 0.8mm}c@{\hskip 0.8mm}c
@{}}
\toprule
&& \multicolumn{3}{c}{\texttt{CNNDM}} & 
\multicolumn{3}{c}{\texttt{XSum}} &
\multicolumn{3}{c}{\texttt{PeerRead}} & \multicolumn{3}{c}{\texttt{PubMed}} & \multicolumn{3}{c}{\texttt{Reddit}} &
\multicolumn{3}{c}{\texttt{AMI}}&
\multicolumn{3}{c}{\texttt{BookSum}} &
\multicolumn{3}{c}{\texttt{MScript}}\\
\cmidrule(lr){3-5} \cmidrule(lr){6-8} \cmidrule(lr){9-11} \cmidrule(lr){12-14} \cmidrule(lr){15-17} 
\cmidrule(lr){18-20} \cmidrule(lr){21-23} \cmidrule(lr){24-26} 
&& \textsc{R} & \textsc{SO} & \textsc{R(P/D/I)}
& \textsc{R} & \textsc{SO} & \textsc{R(P/D/I)}
& \textsc{R} & \textsc{SO} & \textsc{R(P/D/I)}
& \textsc{R} & \textsc{SO} & \textsc{R(P/D/I)}
& \textsc{R} & \textsc{SO} &  \textsc{R(P/D/I)}
& \textsc{R} & \textsc{SO} & \textsc{R(P/D/I)}
& \textsc{R} & \textsc{SO} & \textsc{R(P/D/I)}
& \textsc{R} & \textsc{SO} & \textsc{R(P/D/I)}
\\
\midrule
\parbox[t]{0.8mm}{\multirow{8}{*}{\rotatebox[origin=c]{90}{{\scriptsize{extractive}}}}}

&\textbf{KMeans}   & 
22.2&16.3&14/22/34&
9.8&10.0&14/8/90&
30.9&28.3&24/28/38&
30.6&14.2&31/40/46& 
14.0&12.5&10/2/82&
12.3&2.5&9/6/7&
27.2&4.6&5/2/14&
9.1&0.3&0/0/9
\\
&\textbf{MMR}    & 
21.6&15.2&12/24/30 &
9.8&10.0&14/8/{\color{red}\textbf{97}}&
29.6&24.9&26/29/35&
30.2&12.9&33/35/42& 
13.6&11.5&10/3/{\color{red}\textbf{88}}&
12.3&2.5&9/6/7&
29.1&6.1&4/0/13&
9.5&0.2&0/0/28
\\
&\textbf{TexRank} &  
19.6&10.3&34/27/27&
9.9&8.5&19/11/16&
23.9&12.4&32/32/32&
18.0&1.7&19/21/20&  
17.7&16.7&13/9/15&
11.1&0.0&17/20/6&
6.7&0.0&8/14/8&
8.2&0.2&5/9/8
\\

&\textbf{LexRank} &  
29.3&29.5&71/29/32&
11.2&11.9&{\color{blue}\textbf{61}}/15/19&
29.0&24.6&{\color{blue}\textbf{66}}/35/38&
26.3&7.7&{\color{blue}\textbf{56}}/27/28&  
18.7&18.8&46/11/19&
8.0&0.2&36/21/12&
10.5&0.8&20/20/13&
12.7&\textbf{0.5}&20/9/9%
\\
&\textbf{wILP} &
23.1&15.6&27/28/29&
11.1&2.1&28/19/21&
20.2&16.0&23/27/26&
15.6&6.0&14/20/18& 
17.4&13.5&42/16/20&
5.1&0.6&17/18/17&
4.3&1.3&5/12/7&
6.8&0.1&6/8/6
\\
&\textbf{CL} &  
\textbf{31.2}&30.0&{\color{blue}\textbf{86}}/29/31&%
\textbf{11.8}&\textbf{14.3}&25/13/19&
31.3&21.8&55/35/38&%
26.3&9.2&41/26/26&%
19.4&24.0&23/14/23&%
23.1&10.3&19/23/5&%
-&-&-/-/-&
14.0&0.2&6/8/7%
\\
&\textbf{SumRun}  &  
30.5&27.1&68/29/31&%
11.6&13.1&14/13/19&
\textbf{34.0}&20.5&38/36/37&%
29.4&10.8&27/28/27&%
\textbf{20.2}&19.8&23/12/21&%
\textbf{23.8}&11.4&21/23/6&%
-&-&-/-/-&
14.4&0.0&5/9/9
\\
&\textbf{S2SExt} &  
30.4&28.3&74/28/31&%
12.0&14.2&17/13/19&
33.9&21.1&43/35/37&%
29.6&10.8&26/28/28&%
21.5&\textbf{34.4}&27/12/26&%
23.4&11.9&21/24/6&
-&-&-/-/-&
14.3&0.0&7/9/8%
\\
\midrule
\parbox[t]{0.8mm}{\multirow{4}{*}{\rotatebox[origin=c]{90}{{\scriptsize{abstractive}}}}}
&\textbf{cILP}  &  
27.8&x&43/31/32&
10.9&x&49/15/18&
28.2&x&35/36/38&
27.8&x&23/29/30 &
17.7&x&{\color{blue}\textbf{53}}/15/17 &
12.5&x&22/33/10&
7.9&x&9/19/12&
10.6&x&5/7/7
\\
&\textbf{S2SAbs}&
16.3&x&4/4/4&
10.4&x&8/7/8&
9.9&x&9/9/9&
10.2&x&10/10/10&
11.9&x&11/7/8&
20.3&x&9/12/1&
-&-x&-/-/-&
14.0&x&6/8/8
\\
&+\textbf{Pointer} &
23.9&x&20/13/14&
15.6&x&12/11/12&
13.6&x&13/13/13&
11.2&x&11/12/11&
14.3&x&14/10/12&
23.0&x&11/13/1&
-&-x&-/-/-&
10.0&x&6/7/7
\\
&+\textbf{Teacher} &
29.7&x&33/21/22&
17.0&x&12/10/12&
8.7&x&8/8/8&
11.3&x&12/12/11&
15.3&x&15/10/11&
20.2&x&9/13/1&
-&-x&-/-/-&
16.0&x&7/10/8%
\\
&+\textbf{RL} & 
30.2&x&34/23/24&
18.1&x&12/11/12&
30.1&x&30/29/28&
12.9&x&13/14/13&
16.7&x&1/1/14&
23.6&x&11/13/2&
-&-x&-/-/-&
\textbf{16.2}&x&7/10/8
\\
\midrule
\parbox[t]{0.8mm}{\multirow{4}{*}{\rotatebox[origin=c]{90}{{\scriptsize{ensemble}}}}}
& \textsc{asp}(rand)&
23.3 & 19.5 & 40/{\color{green}\textbf{38}}/{\color{red}\textbf{38}}&
9.0 & 9.0 & 40/{\color{green}\textbf{39}}/38&%
29.6&25.5&54/49/52&
29.5&13.5&49/{\color{green}\textbf{47}}/{\color{red}\textbf{51}}&
12.5&5.2&21/11/22&
8.9 & 0.9 & 44/{\color{green}\textbf{50}}/{\color{red}\textbf{20}}&
\textbf{29.8} & \textbf{6.4} & 57/33/55&
8.4&0.4&{\color{blue}\textbf{32}}/36/37%
\\
& \textsc{asp}(topk)&
29.1 & \textbf{30.4} & 71/31/31&
9.0&8.8&43/{\color{green}\textbf{39}}/38&
30.5&28.2&63/{\color{green}\textbf{54}}/{\color{red}\textbf{57}}&
29.7&14.0&55/48/52&
12.3&15.6&41/{\color{green}\textbf{41}}/38&
9.9 & 1.5 & {\color{blue}\textbf{99}}/24/11&
29.6&6.2&{\color{blue}\textbf{58}}/{\color{green}\textbf{34}}/{\color{red}\textbf{56}}&
8.3&\textbf{0.5}&30/{\color{green}\textbf{37}}/{\color{red}\textbf{38}}%
\\
& \textsc{ext}(rand)&
24.2&20.2&39/25/27&
10.2&10.9&17/13/23&%
29.4&23.5&42/37/39&%
31.7&16.0&37/34/38&%
14.2&17.7&22/12/13&%
18.7&5.1&21/28/8&%
28.6&5.4&37/24/42&%
6.7&0.0&5/9/13%
\\
& \textsc{ext}(topk)&
29.4&30.3&58/25/28&%
11.0&11.8&18/10/37&%
33.0&\textbf{33.0}&54/39/44&%
\textbf{34.1}&\textbf{20.5}&41/35/40&%
16.4&20.8&21/11/52&%
\textbf{23.8}&\textbf{13.4}&23/27/6&%
28.5&5.2&37/24/43&%
7.4&0.0&6/8/11%
\\
\bottomrule
\end{tabularx}

\caption{\label{tab:summarization} Comparison of different systems using the averaged ROUGE scores (1/2/L) with target summaries (\textsc{R}) and averaged oracle overlap ratios (\texttt{SO}, only for extractive systems).
We calculate \textsc{R} between systems and selected summary sentences from each sub-aspect (\textsc{R(P/D/I)}) where each aspect uses the best algorithm: First, ConvexFall and NNearest.
\textsc{R(P/D/I)} is rounded by the decimal point.
- indicates the system has too few samples to train the neural systems.
x indicates \texttt{SO} is not applicable because abstractive systems have no sentence indices.
The best score for each corpora is shown in bold with different colors.
}

\end{table*}

\section{Analysis on System Bias}\label{sec:analysis_system}
We study how current summarization systems are biased with respect to three sub-aspects.
In addition, we show that a simple ensemble of systems shows comparable performance to the single-aspect systems.

\paragraph{Existing systems.}
We compare various extractive and abstractive systems:
For extractive systems, we use \textit{K-Means} \cite{lin2010multi}, Maximal Marginal Relevance (\textit{MMR}) \cite{carbonell1998use}, \textit{cILP} \cite{gillick2009scalable,boudin2015concept}, \textit{TexRank} \cite{mihalcea2004textrank}, \textit{LexRank}~\cite{erkan2004lexrank} and three recent neural systems; \textit{CL} \cite{cheng2016neural}, \textit{SumRun} \cite{nallapati2017summarunner}, and \textit{S2SExt} \cite{kedzie2018content}.
For abstractive systems, we use \textit{WordILP} \cite{banerjeemulti2015ijcai} and four neural systems; \textit{S2SAbs} \cite{rush2015neural}, \textit{Pointer} \cite{see2017get}, \textit{Teacher} \cite{bengio2015scheduled}, and \textit{RL} \cite{paulus2017deep}.
The detailed description and experimental setup for each algorithm are in Appendix.

\paragraph{Proposed ensemble systems.}
\taehee{Should we use ensemble systems are motivated by lin et al? It sounds incorrect.}
Motivated by the sub-aspect theory \cite{lin2012learning,lin2011class}, we combine different types of systems together from two different pools of extractive systems: \textsc{asp} from the three best algorithm from each aspect and \textsc{ext} from all extractive systems.
For each combination, we choose the sumary sentences randomly among the union set of the predicted sentences (rand) or the most frequent unique sentences (topk).

\paragraph{Results.}
Table \ref{tab:summarization} shows a comparison of existing and proposed summarization systems on the set of corpora in \S\ref{sec:data} except for \texttt{Newsroom}\footnote{We exclude it because of its similar behavior as \texttt{CNNDM}.}.
Neural extractive systems such as \textit{CL}, \textit{SumRun} and \textit{S2SExt} outperform the others in general. 
\textit{LexRank} is highly biased toward the position aspect. 
On the other hand, \textit{MMR} is extremely biased to the importance aspect on \texttt{XSum} and \texttt{Reddit}.
Interestingly, neural extractive systems are somewhat balanced compared to the others.
Ensemble systems seem to have the three sub-aspects in balance, compared to the neural extractive systems. 
They also outperform the others (either \texttt{ROUGE} or \texttt{SO}) on five out of eight datasets. 


\section{Conclusion and Future Directions}\label{sec:conclusion}

We define three sub-aspects of text summarization: position, diversity, and importance. We analyze how different domains of summarization dataset are biased to these aspects. 
We observe that news articles strongly reflect the position aspect, while the others do not.
In addition, we investigate how current summarization systems reflect these three sub-aspects in balance. 
Each type of approach has its own bias, while neural systems rarely do.
Simple ensembling of the systems shows more balanced and comparable performance than single ones.

We summarize actionable messages for future summarization research:

\begin{itemize}[noitemsep,topsep=0pt,leftmargin=*]
\item Different domains of datasets except for news articles pose new challenges to the appropriate design of summarization systems.
For example, summarization of conversations (e.g., \texttt{AMI}) or dialogues (\texttt{MSCript}) need to filter out repeated, rhetorical utterances. 
Book summarization (e.g., \texttt{BookSum}) is very challenging due to its extremely large document size.  Here current neural encoders suffer from computation limits.
\item Summarization systems to be developed should clearly state their computational limits as well as effectiveness in each aspect and in each corpus domain.
A good summarization system should reflect different kinds of the sub-aspects harmoniously, regardless of corpus bias. Developing such bias-free or robust models can be very important for future directions.
\item Nobody has clearly defined the deeper nature of meaning abstraction yet. A more theoretical study of summarization, and the various aspects, is required. A recent notable example is \citet{peyrard-2019-simple}'s attempt to theoretically define different quantities of \texttt{importance} aspect, and demonstrate the potential of the framework on an existing summarization system. Similar studies can be applied to other aspects and their combinations in various systems and different domains of corpora.
\item One can repeat our bias study on evaluation metrics.
\citet{peyrard-2019-studying} showed that widely used evaluation metrics (e.g., ROUGE, Jensen-Shannon divergence) are strongly mismatched in scoring summary results. 
One can compare different measures (e.g., n-gram recall, sentence overlaps, embedding similarities, word connectedness, centrality, importance reflected by discourse structures), and study bias of each with respect to systems and corpora.
\end{itemize}


\section*{Acknowledgements}
This work would not have been possible without the efforts of the authors who kindly share the summarization datasets publicly. 
We thank Rada Mihalcea for sharing the book summarization dataset.
We also thank Diane J. Litman, Taylor Berg-Kirkpatrick, Hiroaki Hayashi, and anonymous reviewers for their helpful comments.

\bibliographystyle{acl_natbib}
\bibliography{neuralsum}

\clearpage

\renewcommand*\appendixpagename{\Large Appendices}
\begin{appendix}\label{sec:appendix_dialrec}

\section{Systems and Setup: Details}

For extractive systems,
\textit{K-Means} rank sentences clusters by descending order of cluster sizes, and then using a greedy algorithm \cite{lin2010multi} to select the nearest sentences to the centroid.
Maximal Marginal Relevance (\textit{MMR}) finds sentences which are highly relevant to the document but less redundant with sentences already selected for a summary.
\textit{cILP} \cite{gillick2009scalable,boudin2015concept} weights sub-sentences and maximizes their coverage by minimizing redundancy globally using Integer Linear Program (ILP).
\textit{TexRank} \cite{mihalcea2004textrank} automatically extracts keywords using Levenshtein distance between the text keywords.
\textit{LexRank}~\cite{erkan2004lexrank} uses module centrality for ranking the keywords.
In addition, we also use the recent three neural extractive systems: \textit{CL} \cite{cheng2016neural}, \textit{SumRun} \cite{nallapati2017summarunner}, and \textit{S2SExt} \cite{kedzie2018content}, where each has a little variation in their extraction architecture\footnote{See  \cite{kedzie2018content} for a detailed comparison.}.

In training \textit{CL}, \textit{SumRun}, and \textit{S2SExt}, we use upweight positive labels to make them proportional to the negative labels. 
We use 200 embedding size of GloVe \cite{pennington2014glove} pre-trained embeddings with 0.25 dropout on embeddings, fixing it not to be trained during training.
We use CNN encoder with 6 window size as [25, 25, 50, 50, 50, 50] feature maps.
We use 1-layer of sequence-to-sequence model with 300 size of LSTM and 100 size of MLP with 0.25 dropout.
\textit{SumRun} uses 16 size of segment and 16 size of position embeddings.

For abstractive systems, we use \textit{WordILP} \cite{banerjeemulti2015ijcai} that produces a word graph of important sentences and then choose sentences from the word graph employing a ILP solver.
We also use incremental sequence-to-sequence models: a basic \textit{S2SAbs} \cite{rush2015neural} with \textit{Pointer} network \cite{see2017get}, with teacher forcing \textit{Teacher} \cite{bengio2015scheduled}, and with reinforcement learning on the evaluation metrics, and \textit{RL} \cite{paulus2017deep}.

In training \textit{S2SAbs}, \textit{Pointer}, \textit{Pointer}, and \textit{RL}, 
we use 150 hidden size of GRU with 300 size of GloVe embeddings.
\textit{Pointer} uses maximum coverage function using NLL loss. 
\textit{Teacher} uses 0.75 ratio of teach forcing with exponential decaying function.
and \textit{RL} uses 0.1 ratio of RL optimization after the first epoch of \textit{S2SAbs} training.
We use 4 size of beam searching at decoding.
We use 32 batch size with adam optimizer of 0.001 learning rate. 

For \texttt{MScript}, the original dataset has no data split, so we randomly split it by 0.9, 0.05, 0.05 for train, valid, test set, respectively.

\section{Venn Diagram for All Datasets}

\begin{figure*}[h]
\vspace{0mm}
\centering
{
\subfloat[CNNDM(49.4\%)]{
\includegraphics[trim=0cm 0cm 0cm 1cm,clip,valign=b,width=.33\linewidth]{figs/venndiagram/venndiagram_cnndm_test_first-knn-hard_convex_waterfall.png}}
\subfloat[Newsr.(54.4\%)]{
\includegraphics[trim=0cm 0cm 0cm 1cm,clip,valign=b,width=.33\linewidth]{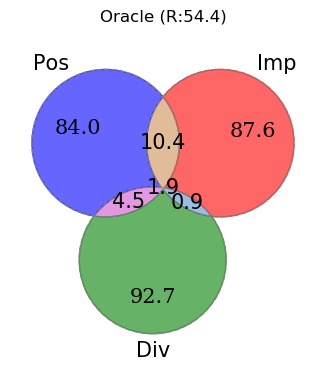}}
\subfloat[XSum(76.8\%)]{
\includegraphics[trim=0cm 0cm 0cm 1cm,clip,valign=b,width=.33\linewidth]{figs/venndiagram/venndiagram_xsum_clean_test_first-knn-hard_convex_waterfall.png}}
\\
\subfloat[PeerRead(37.64\%)]{
\includegraphics[trim=0cm 0cm 0cm 1cm,clip,valign=b,width=.33\linewidth]{figs/venndiagram/venndiagram_peerread_test_first-knn-hard_convex_waterfall.png}}
\subfloat[PubMed(64.0\%)]{
\includegraphics[trim=0cm 0cm 0cm 1cm,clip,valign=b,width=.33\linewidth]{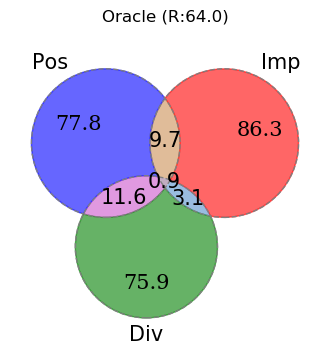}}
\subfloat[Reddit(68.1\%)]{
\includegraphics[trim=0cm 0cm 0cm 1cm,clip,valign=b,width=.33\linewidth]{figs/venndiagram/venndiagram_reddit_all_first-knn-hard_convex_waterfall.png}}
\\
\subfloat[AMI(94.1\%)]{
\includegraphics[trim=0cm 0cm 0cm 1cm,clip,valign=b,width=.33\linewidth]{figs/venndiagram/venndiagram_ami_all_first-knn-hard_convex_waterfall.png}}
\subfloat[BookSum(87.1\%)]{
\includegraphics[trim=0cm 0cm 0cm 1cm,clip,valign=b,width=.33\linewidth]{figs/venndiagram/venndiagram_booksum_test_first-knn-hard_convex_waterfall.png}}
\subfloat[MScript (99.1\%)]{
\includegraphics[trim=0cm 0cm 0cm 1cm,clip,valign=b,width=.33\linewidth]{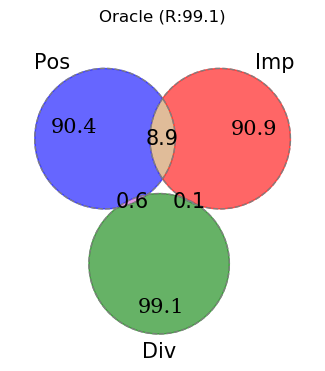}}
}
\caption{\label{fig:venndiagram_appendix} Venndiagram of averaged summary sentence overlaps across the the sub-aspects for all datasets.
We use First-k for \textsc{Position} (P), ConvexFall for \textsc{Diversity} (D), and N-Nearest for \textsc{Importance} (I).
The number called \textit{Oracle Recall} in the parenthesis is the averaged ratio of how many the oracle sentences are NOT chosen by union set of the three sub-aspect algorithms.
}
\vspace{0mm}
\end{figure*}

Sentence Venn diagrams among three aspects and oracle for all datasets are shown in Figure~\ref{fig:venndiagram_appendix}. \texttt{Newsroom} has an analogous pattern to \texttt{XSum}. Compared to \texttt{PeerRead}, , \texttt{PubMed} has relatively less sentence overlap between \textsc{First-k} and the other two aspects. \texttt{MScript} has extremely small oracle sentence overlaps to all three aspects. However, it is mainly because of the characteristics of the dataset: it has long source documents (1k sentences on average) with short (5 sentences on average) summary.

\section{Full ROUGE F Scores for Corpus Bias Analysis}

In Table~\ref{tab:analysis_multi_appendix}, we provide a full list of ROUGE F scores for all datasets w.r.t three sub-aspects. We find that in \texttt{MScript}, the best algorithms for each of ROUGE-1/2/L are different. 

\section{Documents in an Embedding Space: for All Datasets}

\begin{figure*}[ht!]
\centering
{
\subfloat[\footnotesize{CNNDM}]{
\includegraphics[trim=0.6cm 1.3cm 0.6cm 2.2cm,clip,width=.44\textwidth]{figs/sample/example_cnndm_8936.png}
\qquad
\includegraphics[trim=0.6cm 1.3cm 0.6cm 2.2cm,clip,width=.44\textwidth]{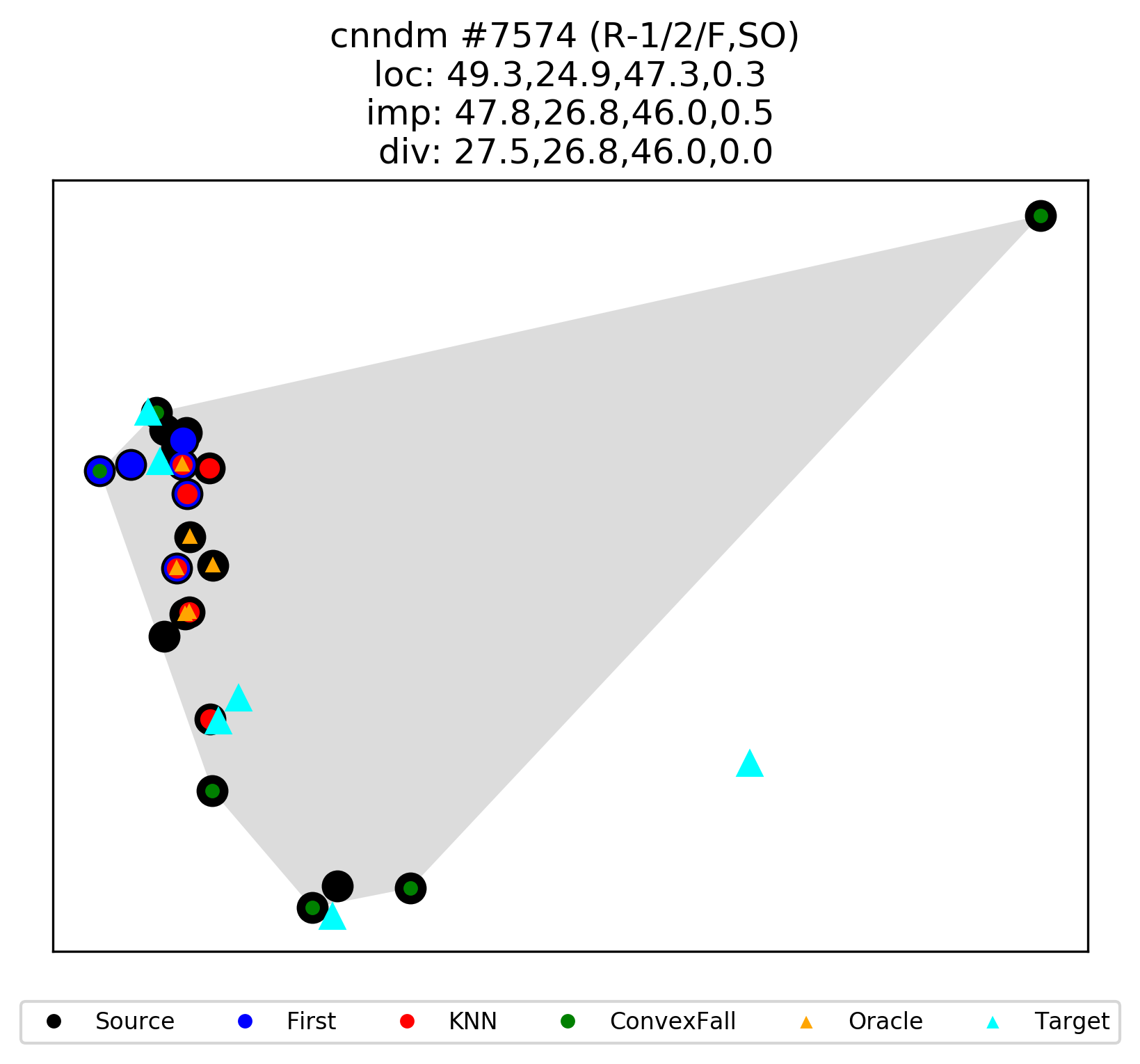}
}
\\
\subfloat[\footnotesize{NewsRoom}]{
\includegraphics[trim=0.6cm 1.3cm 0.6cm 2.2cm,clip,width=.44\textwidth]{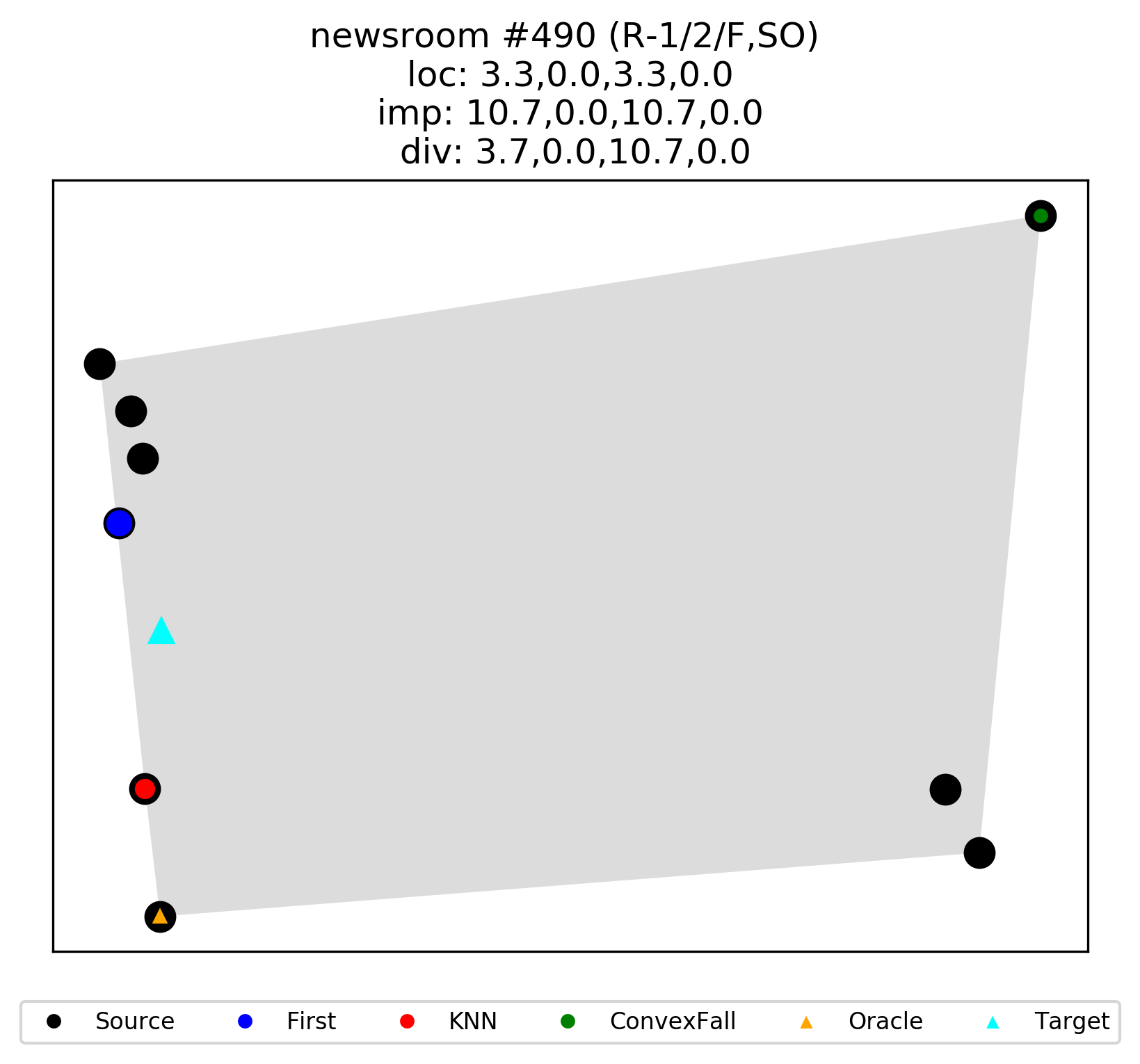}
\qquad
\includegraphics[trim=0.6cm 1.3cm 0.6cm 2.2cm,clip,width=.44\textwidth]{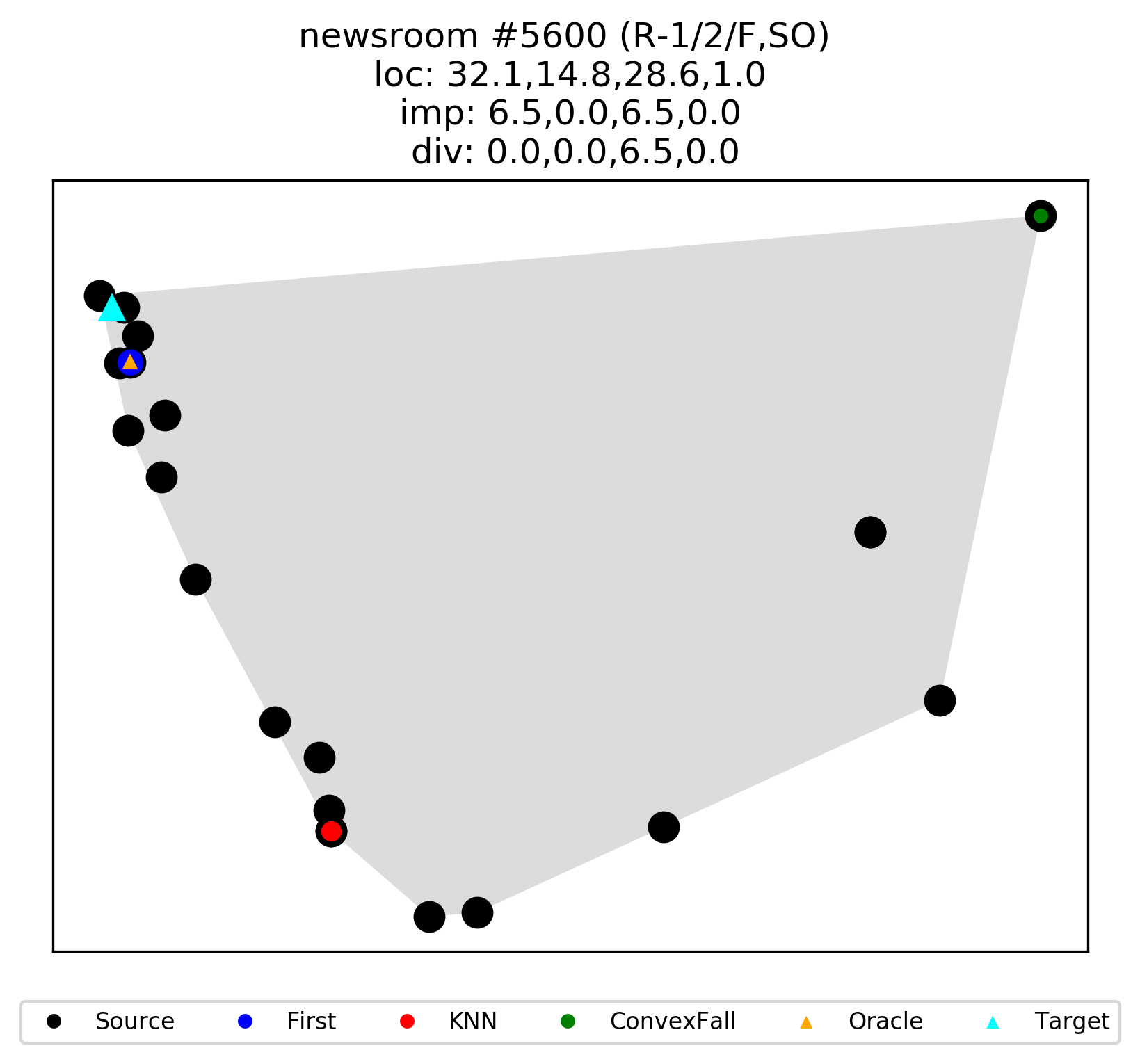}
}
\\
\subfloat[\footnotesize{XSum}]{
\includegraphics[trim=0.6cm 1.3cm 0.6cm 2.2cm,clip,width=.44\textwidth]{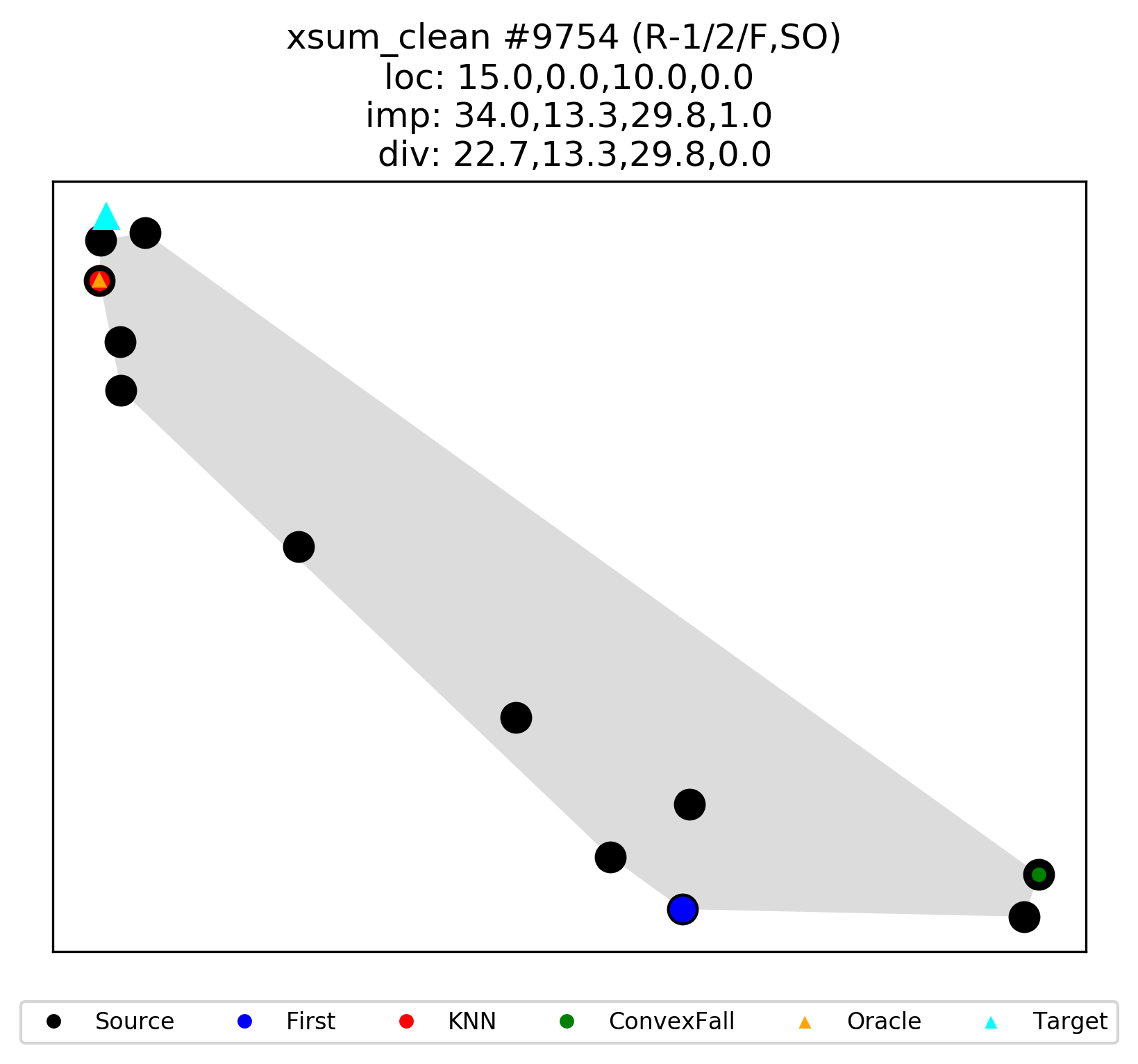}
\qquad
\includegraphics[trim=0.6cm 1.3cm 0.6cm 2.2cm,clip,width=.44\textwidth]{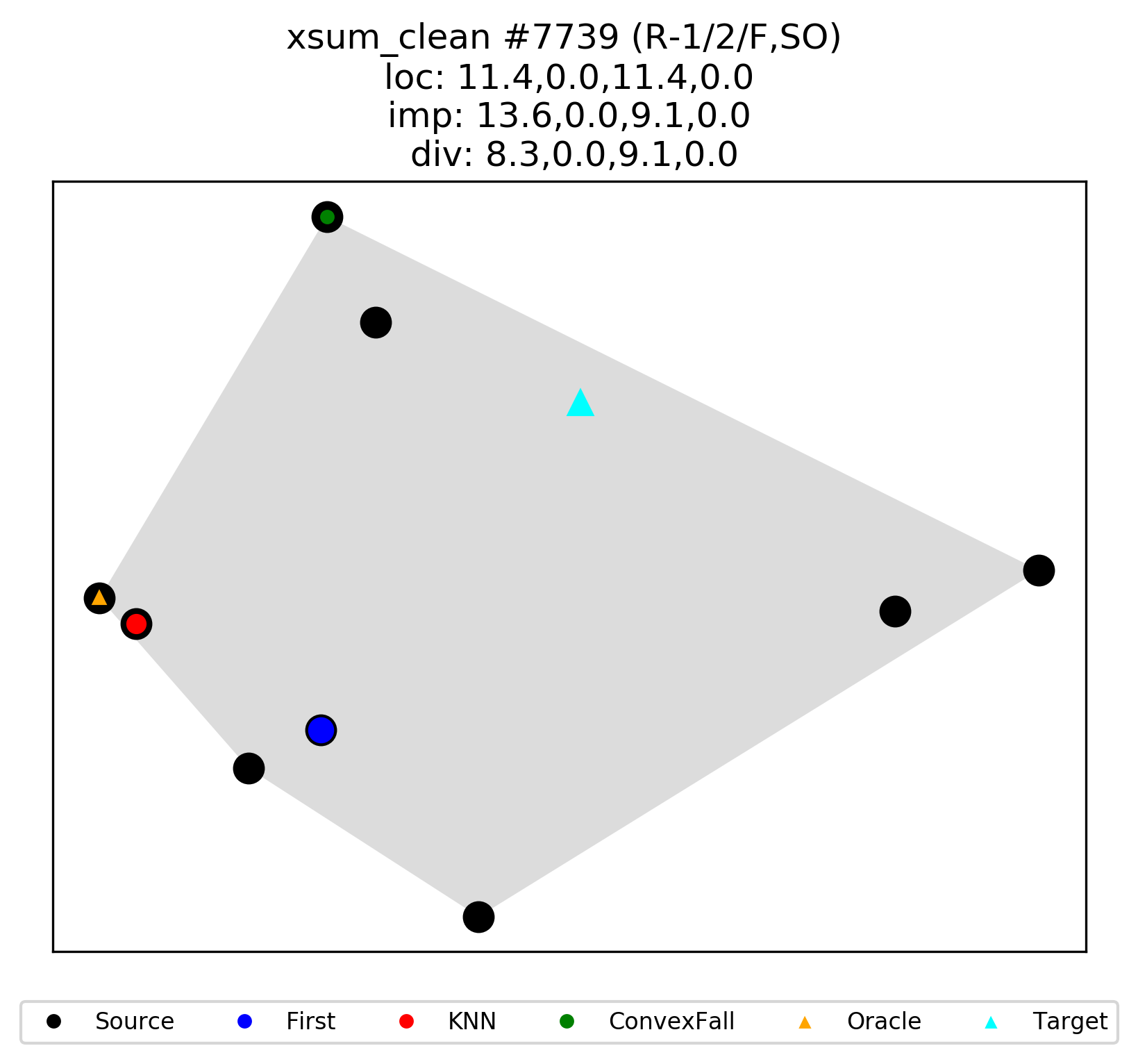}}
\\
\subfloat[\footnotesize{PeerRead}]{
\includegraphics[trim=0.6cm 1.3cm 0.6cm 2.2cm,clip,width=.44\textwidth]{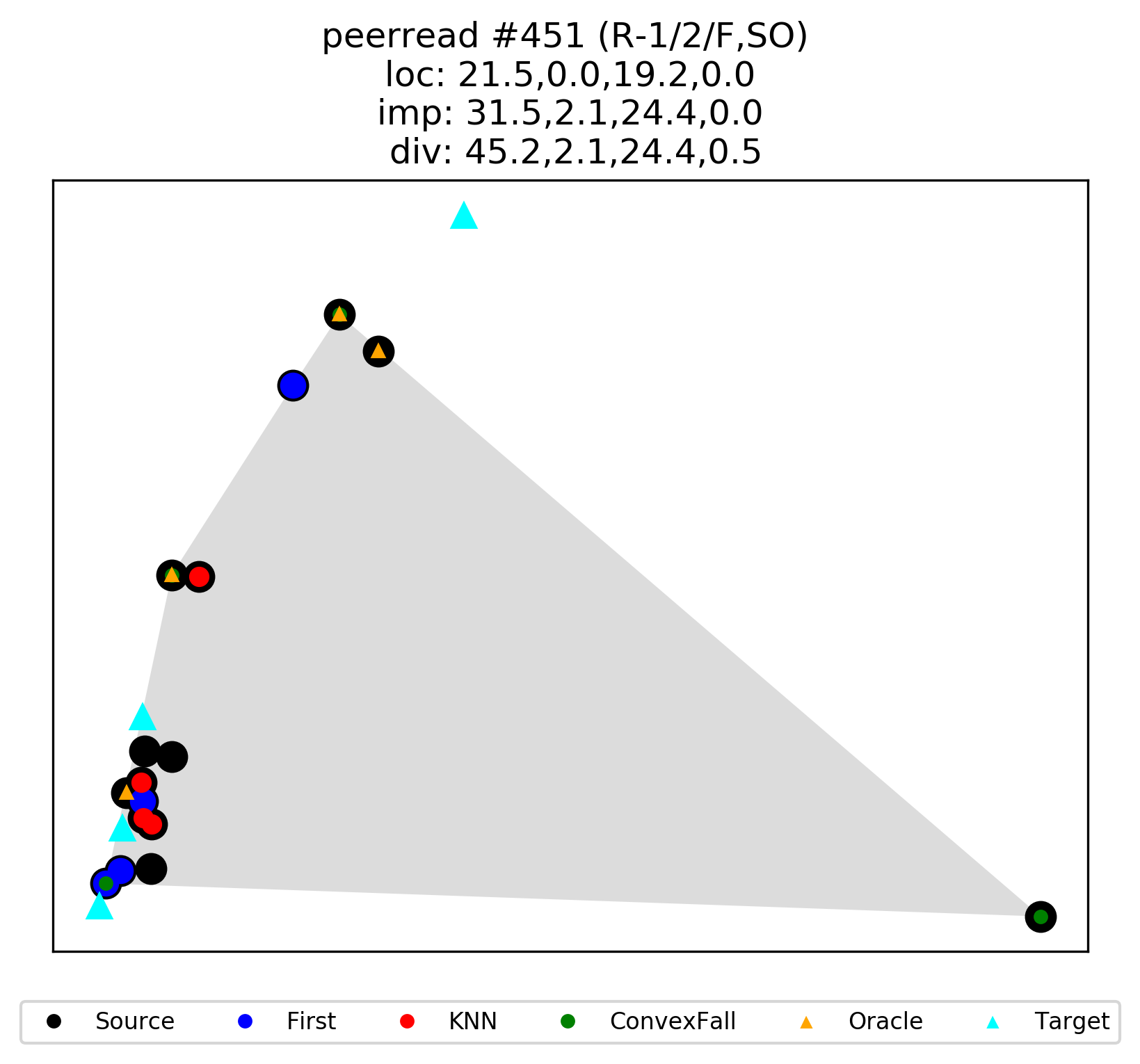}
\qquad
\includegraphics[trim=0.6cm 1.3cm 0.6cm 2.2cm,clip,width=.44\textwidth]{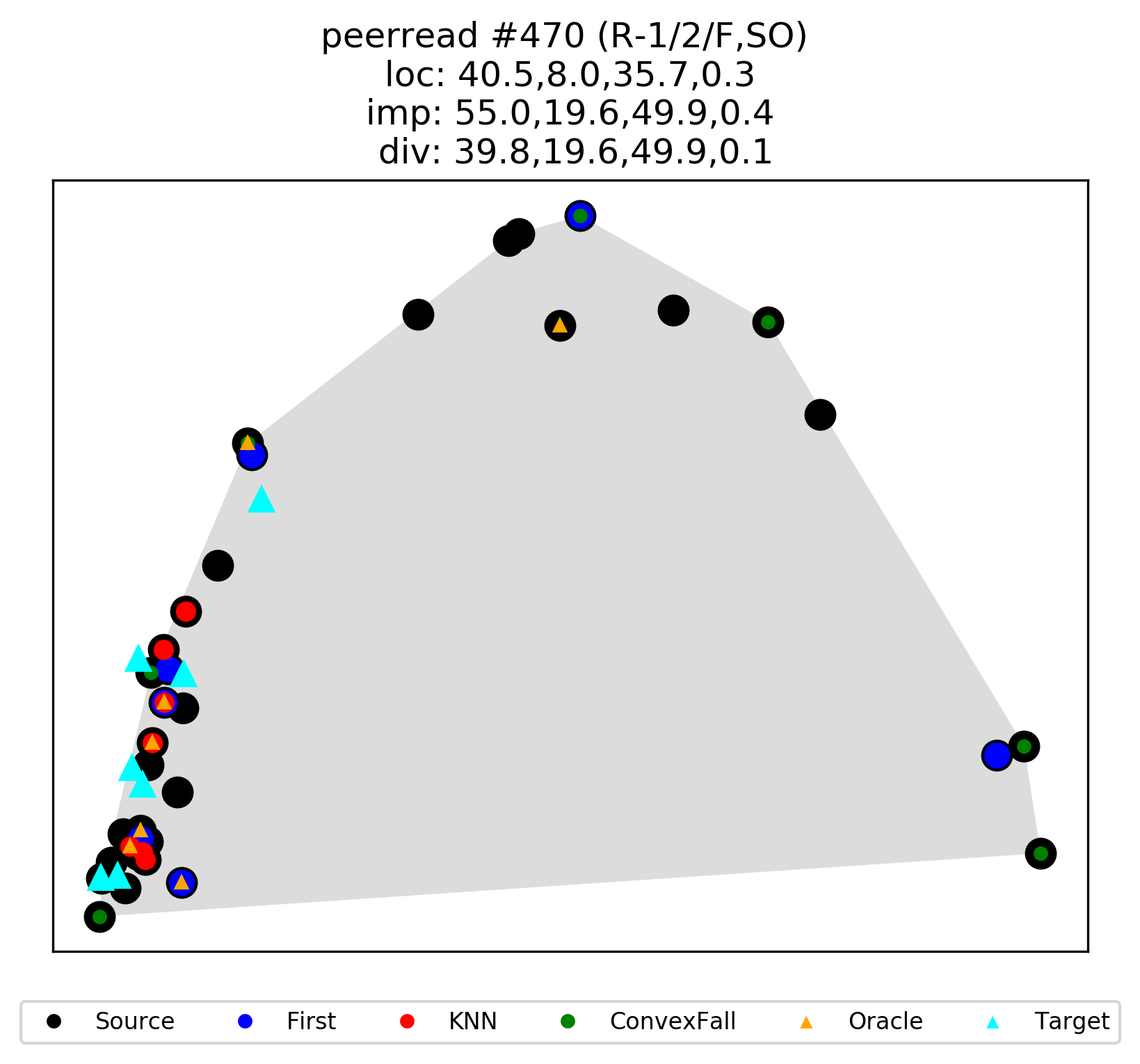}}
}
\caption{\label{fig:pca_multi_aspect1} PCA projection of extractive summaries chosen by multiple aspects of algorithms (\texttt{CNNDM, NewsRoom, XSum, PeerRead}, and \texttt{PubMed}). Source and target sentences are black circles ($\newmoon$) and {\color{purple} purple} stars, respectively. The {\color{blue}blue}, {\color{green}green}, {\color{red}red} circles are summary sentences chosen by \texttt{First}, \texttt{ConvexFall}, \texttt{KN}, respectively.
The {\color{yellow}yellow} stars are the oracle sentences. Best viewed in color.
}
\end{figure*}

\begin{figure*}[ht!]
\centering
\subfloat[\footnotesize{PubMed}]{
\includegraphics[trim=0.6cm 1.3cm 0.6cm 2.2cm,clip,width=.44\textwidth]{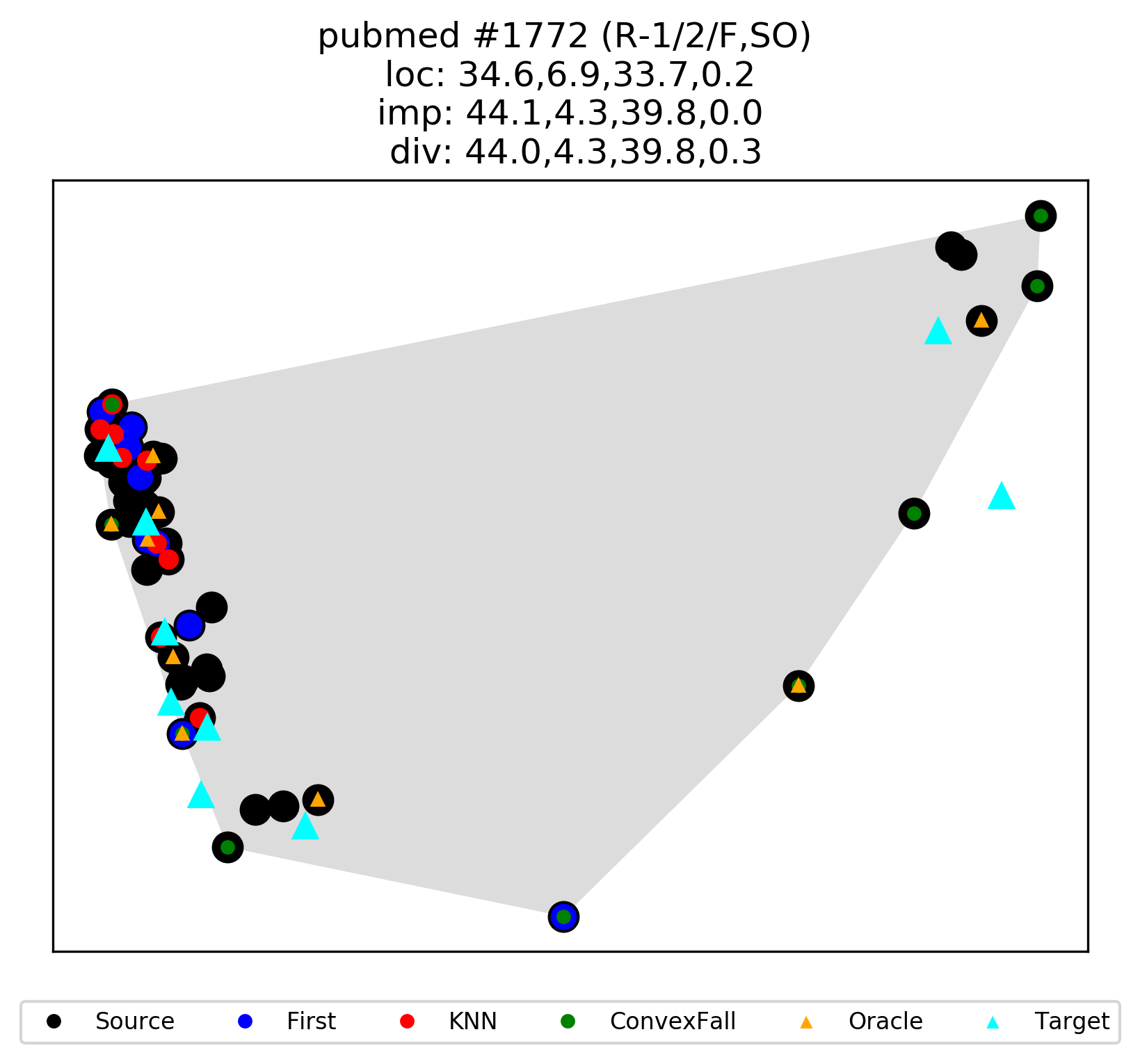}
\qquad
\includegraphics[trim=0.6cm 1.3cm 0.6cm 2.2cm,clip,width=.44\textwidth]{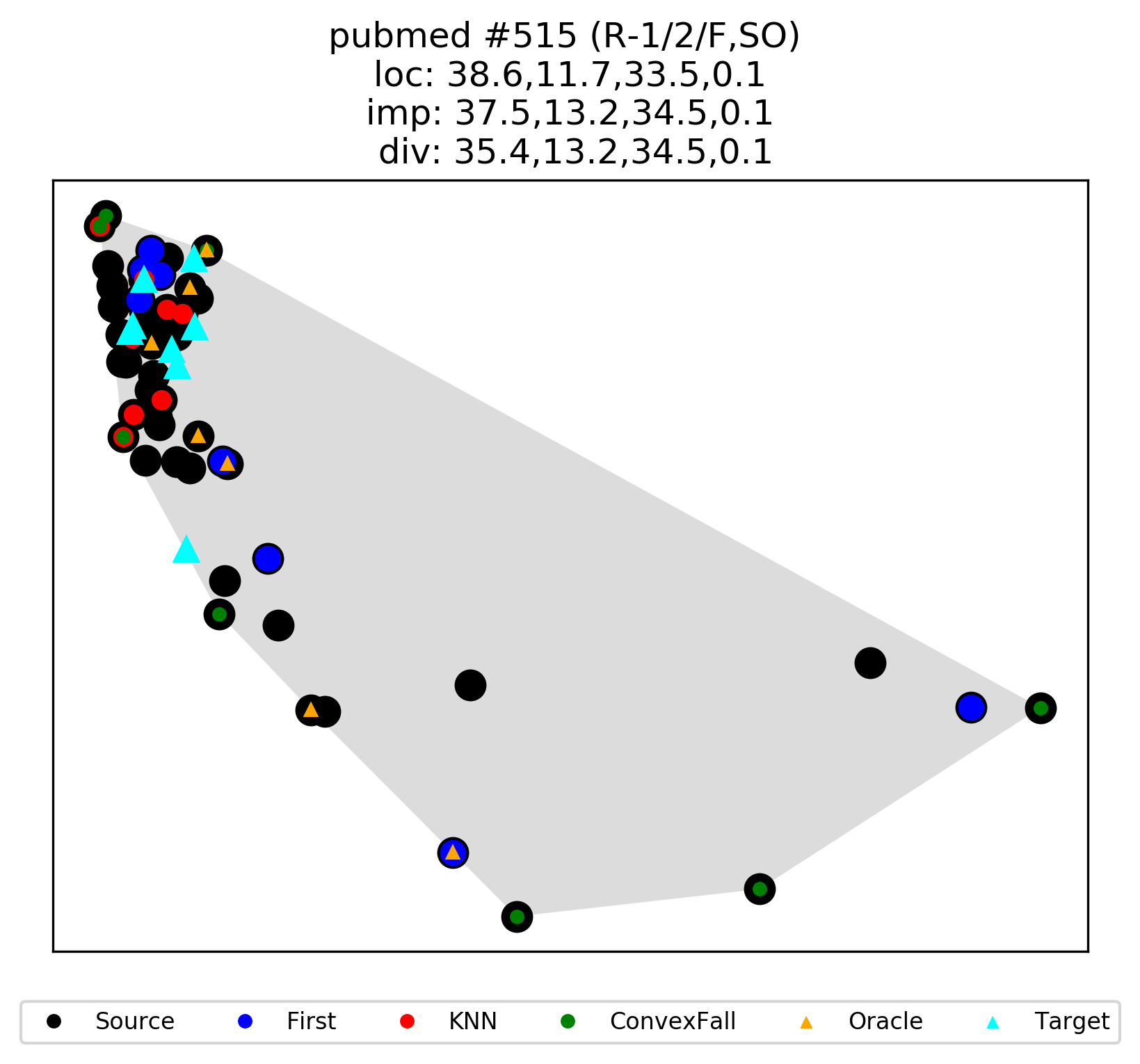}}
\\
\subfloat[\footnotesize{Reddit}]{
\includegraphics[trim=0.6cm 1.3cm 0.6cm 2.2cm,clip,width=.44\textwidth]{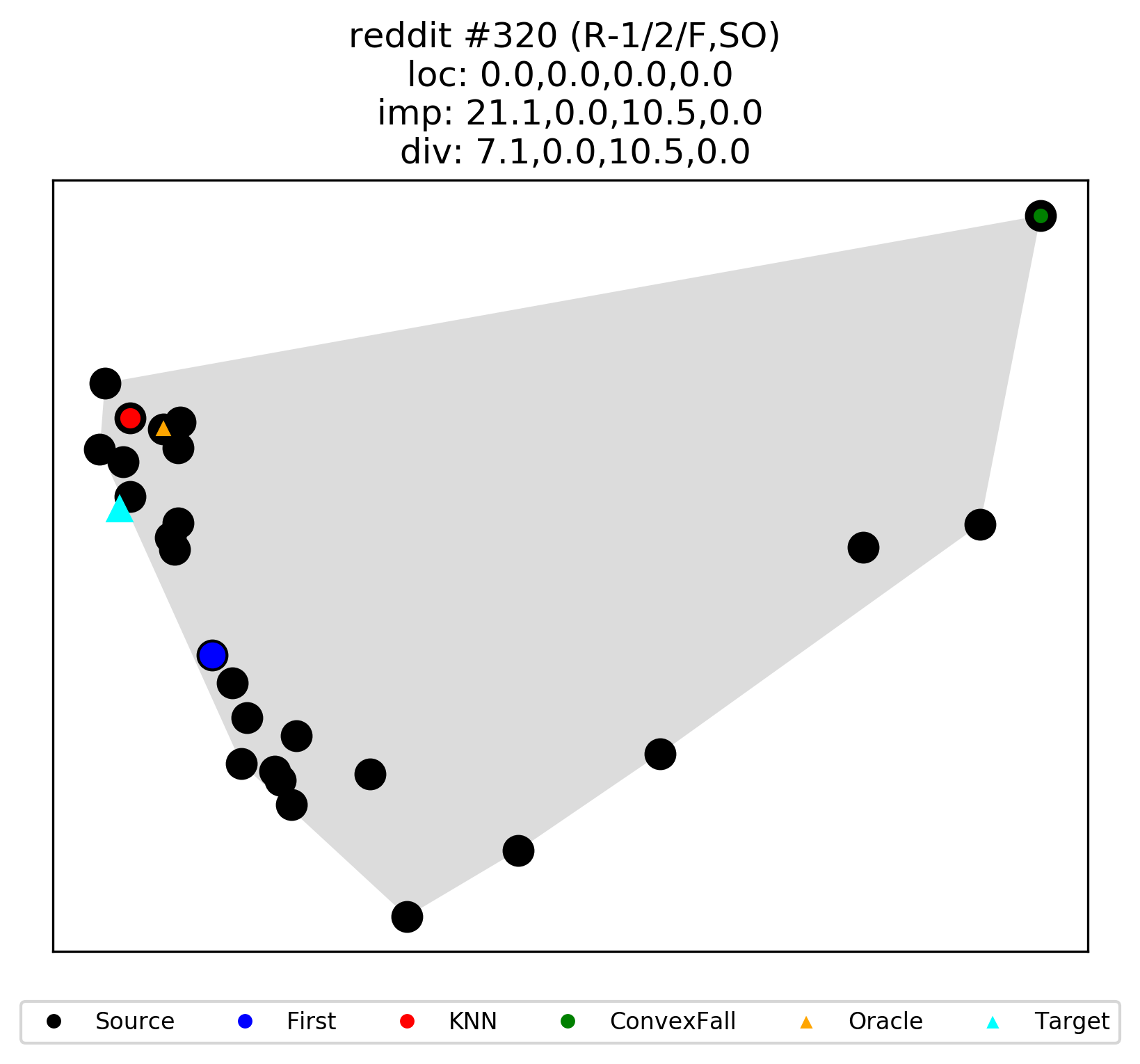}
\qquad
\includegraphics[trim=0.6cm 1.3cm 0.6cm 2.2cm,clip,width=.44\textwidth]{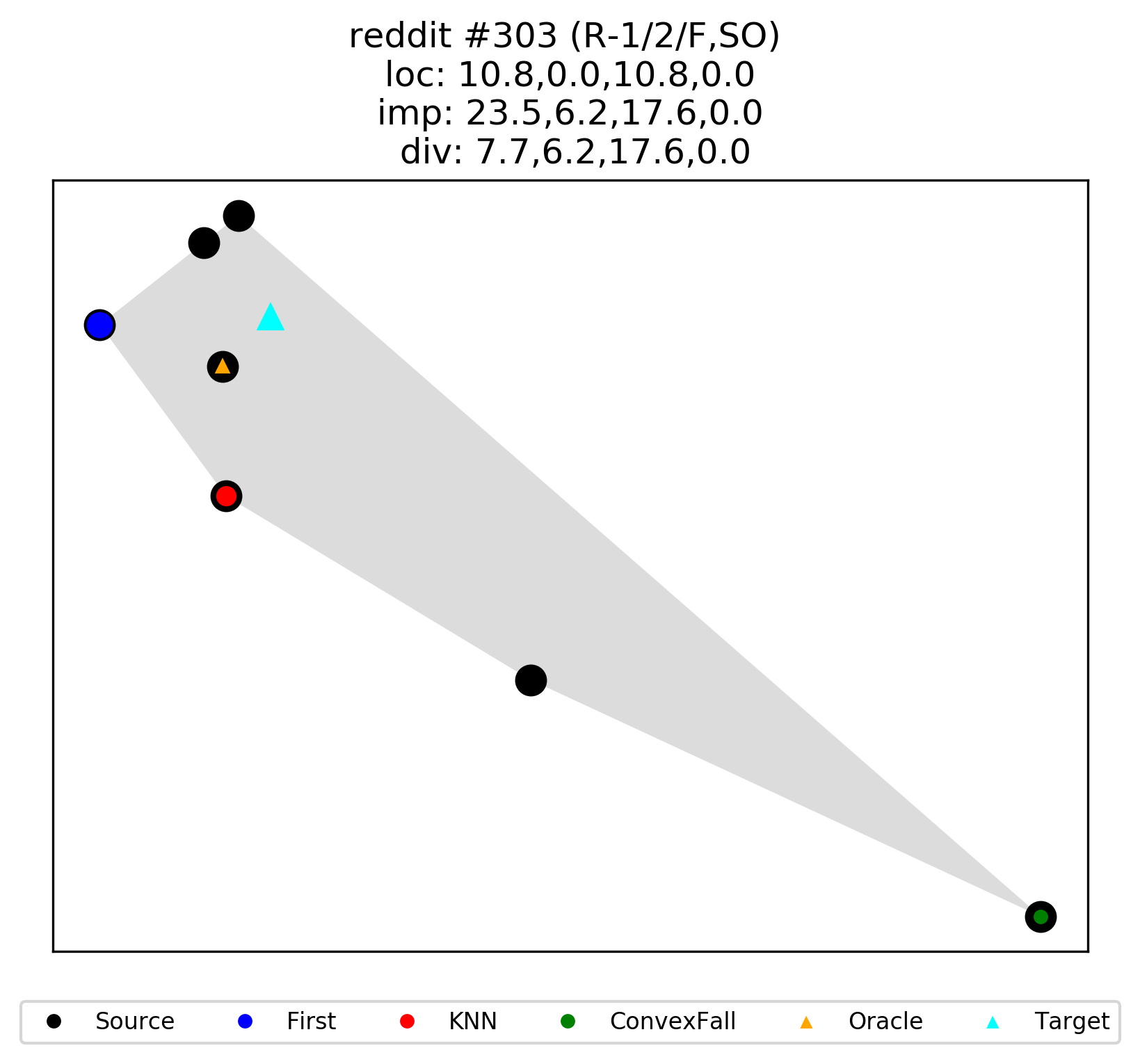}}
\\
\subfloat[\footnotesize{AMI}]{
\includegraphics[trim=0.6cm 1.3cm 0.6cm 2.2cm,clip,width=.44\textwidth]{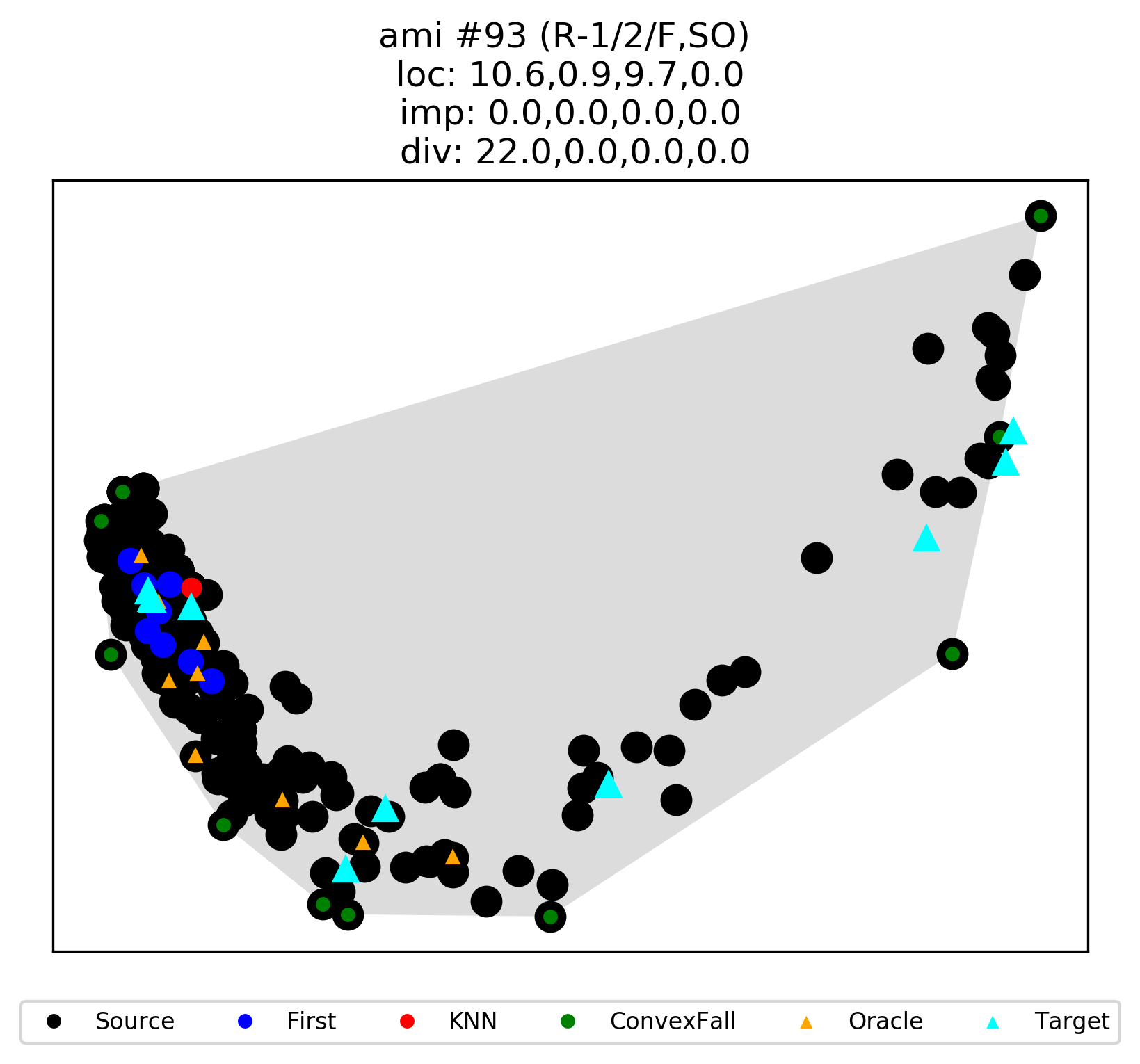}
\qquad
\includegraphics[trim=0.6cm 1.3cm 0.6cm 2.2cm,clip,width=.44\textwidth]{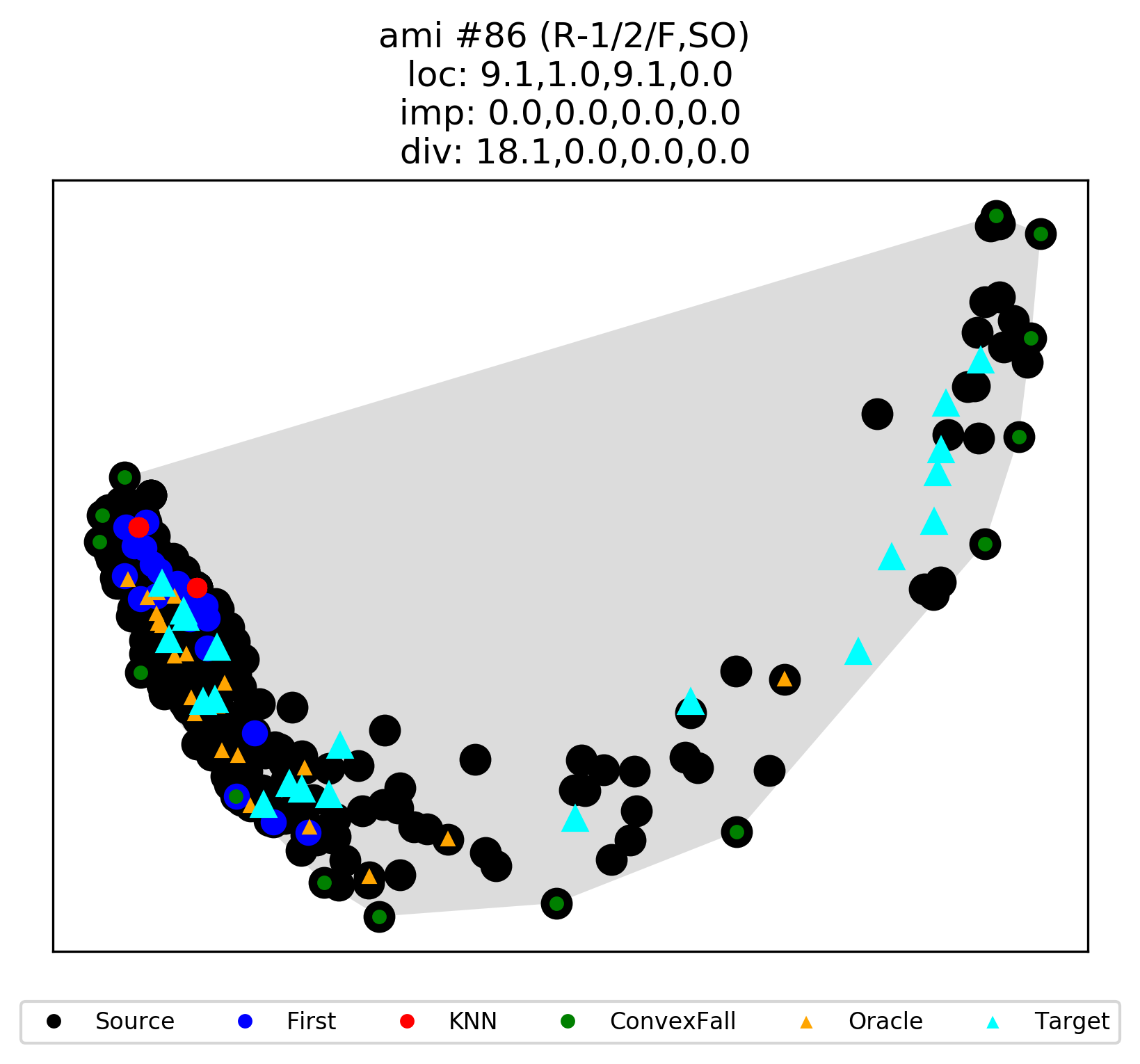}}
\\
\subfloat[\footnotesize{BookSum}]{
\includegraphics[trim=0.6cm 1.3cm 0.6cm 2.2cm,clip,width=.44\textwidth]{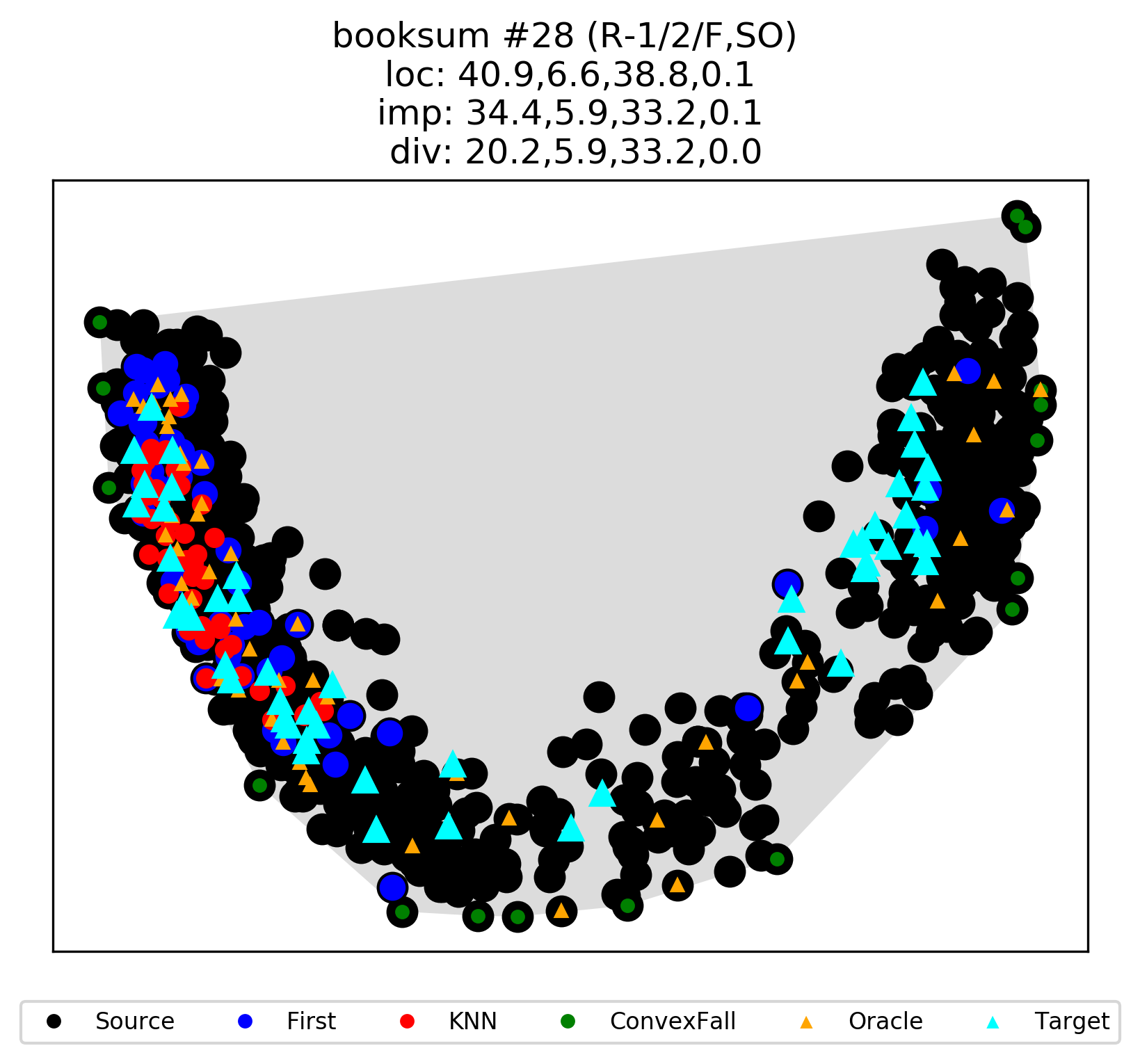}
\qquad
\includegraphics[trim=0.6cm 1.3cm 0.6cm 2.2cm,clip,width=.44\textwidth]{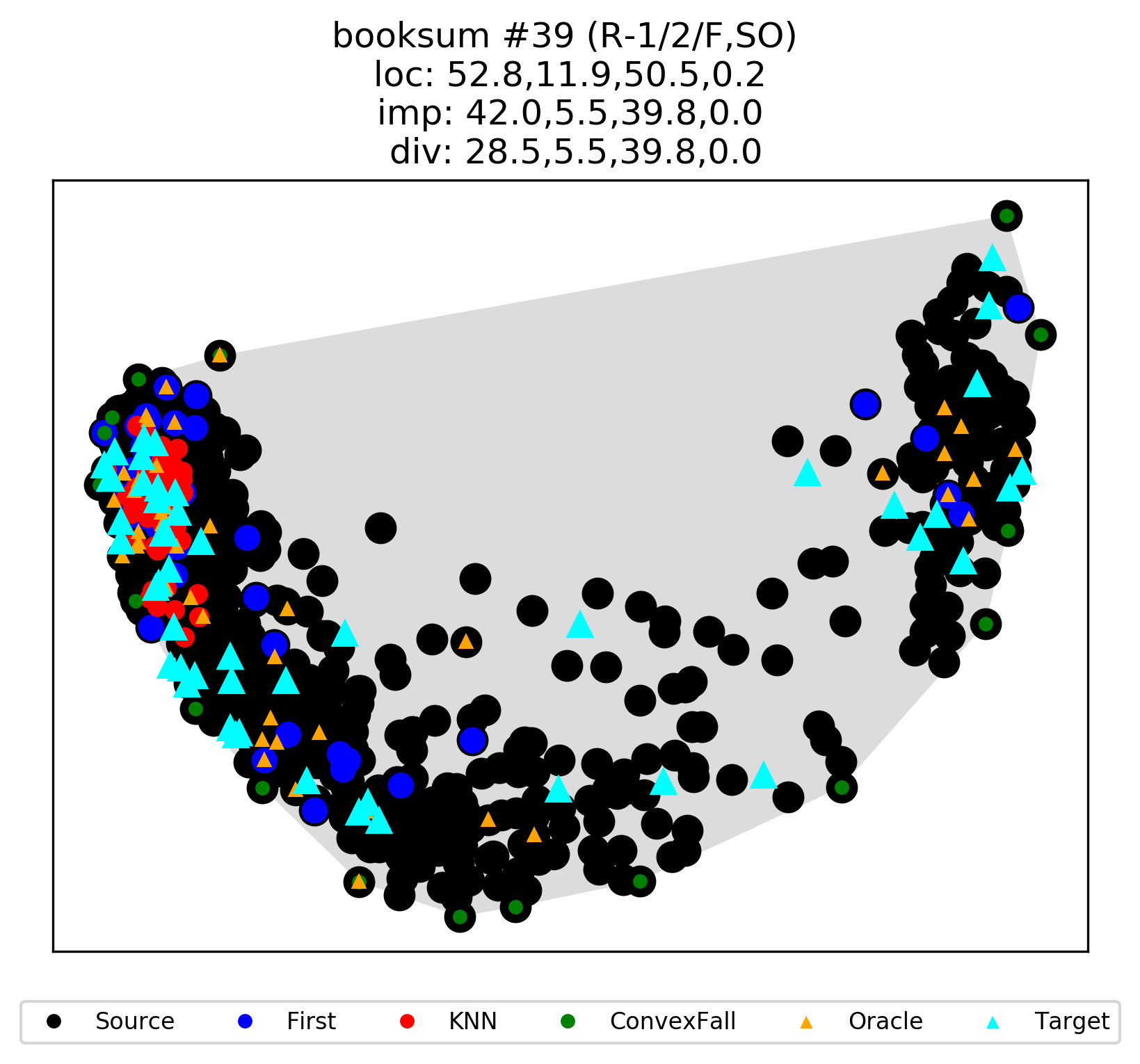}}
\caption{\label{fig:pca_multi_aspect2} PCA projection of extractive summaries chosen by multiple aspects of algorithms (\texttt{Reddit, AMI, Booksum}, and \texttt{MScript}). Source and target sentences are black circles ($\newmoon$) and {\color{purple} purple} stars, respectively. The {\color{blue}blue}, {\color{green}green}, {\color{red}red} circles are summary sentences chosen by \texttt{First}, \texttt{ConvexFall}, \texttt{KN}, respectively.
The {\color{yellow}yellow} stars are the oracle sentences. Best viewed in color.
}
\end{figure*}

In Figure (\ref{fig:pca_multi_aspect1},\ref{fig:pca_multi_aspect2}), we have more two-dimensional PCA projection examples for source documents from all datasets. We find a weak pattern about where target sentences lie on according to the number of them. For example, from \texttt{XSum} and \texttt{Reddit} which have a single target sentence,  we investigate that some target sentences are located in the middle of \texttt{ConvexHull}, which are far from any source sentences. 

\section{System Biases per each corpus with the Three Sub-aspects}

\begin{table}[ht!]
\centering
\footnotesize
\vspace{0mm}
\begin{tabularx}{.99\linewidth}{@{}c@{\hskip 1mm}|@{\hskip 0mm}r@{\hskip 0mm}|
@{\hskip 3mm} c@{\hskip 1mm}c@{\hskip 1mm}c  
@{\hskip 3mm} c@{\hskip 1mm}c@{\hskip 1mm}c 
@{\hskip 3mm} c@{\hskip 1mm}c@{\hskip 1mm}c 
@{\hskip 3mm} c@{\hskip 1mm}c@{\hskip 1mm}c@{}}
\toprule
&& \multicolumn{1}{c}{\texttt{CNNDM}} &  \multicolumn{1}{c}{\texttt{NewsRoom}} & \multicolumn{1}{c}{\texttt{XSum}} \\
\midrule
&& R-1/2/L 
& R-1/2/L
& R-1/2/L \\
\midrule
&\textsc{Random}
&26.6/6.7/23.9
&15.2/2.8/12.2
&14.9/1.8/11.2\\
&\textsc{Oracle} 
&51.5/28.5/48.6
&53.4/40.2/50.7
&27.9/7.5/23.2\\
\midrule
\parbox[t]{1mm}{\multirow{3}{*}{\rotatebox[origin=c]{90}{{\scriptsize{\textsc{Position}}}}}} &
\textsc{First-k}
&\textbf{39.1/17.1/35.8}
&\textbf{36.9/25.9/33.9}
&14.8/1.4/11.1\\
&\textsc{Last-k}
&23.5/4.7/21.1
&11.5/2.0/9.5
&13.2/1.5/10.1\\
&\textsc{Middle-k}
&29.4/8.6/26.4
&17.4/5.3/14.4
&14.7/1.7/11.0\\
\midrule
\parbox[t]{1mm}{\multirow{2}{*}{\rotatebox[origin=c]{90}{{\scriptsize{\textsc{Divers.}}}}}} 
&\textsc{ConvexFall}
&29.5/8.6/26.6
&15.0/4.0/12.7
&13.6/1.3/10.5\\
&\textsc{Heuristic}
&29.2/8.7/26.3
&14.9/4.1/12.7
&13.6/1.3/10.5\\
\midrule
\parbox[t]{1mm}{\multirow{2}{*}{\rotatebox[origin=c]{90}{{\scriptsize{\textsc{Import.}}}}}}
&\textsc{N-Nearest}
&29.7/9.3/26.9
&18.9/6.1/15.7
&\textbf{15.7/2.0/11.7}\\
&\textsc{K-Nearest}
&30.6/10.5/27.8
&19.1/6.8/16.0
&15.0/1.8/11\\
\bottomrule
\end{tabularx}

\vspace{5mm}

\begin{tabularx}{.99\linewidth}{@{}c@{\hskip 1mm}|@{\hskip 0mm}r@{\hskip 0mm}|
@{\hskip 3mm} c@{\hskip 1mm}c@{\hskip 1mm}c  
@{\hskip 3mm} c@{\hskip 1mm}c@{\hskip 1mm}c 
@{\hskip 3mm} c@{\hskip 1mm}c@{\hskip 1mm}c 
@{\hskip 3mm} c@{\hskip 1mm}c@{\hskip 1mm}c@{}}
\toprule
&& \multicolumn{1}{c}{\texttt{PeerRead}}
& \multicolumn{1}{c}{\texttt{PubMed}} 
&\multicolumn{1}{c}{\texttt{Reddit}}\\
\midrule
&& R-1/2/L 
& R-1/2/L
& R-1/2/L \\
\midrule
&\textsc{Random}
&38.2/11.1/34.3
&41.3/11.3/37.6
&17.6/3.7/14.2
\\
&\textsc{Oracle} 
&56.6/29.5/52.7
&58.2/27.9/54.8
&38.5/17.8/33.8
\\
\midrule
\parbox[t]{1mm}{\multirow{3}{*}{\rotatebox[origin=c]{90}{{\scriptsize{\textsc{Position}}}}}} &
\textsc{First-k}
&\textbf{41.4/16.8/37.9}
&37.8/10.2/34.7
&\textbf{21.8/6.2/17.8}
\\
&\textsc{Last-k}
&39.1/12.4/35.1
&39.1/11.8/35.9
&116.4/3.7/13.4
\\
&\textsc{Middle-k}
&40.4/12.5/36.3
&39.5/10.8/36.3
&17.4/3.2/13.8
\\
\midrule
\parbox[t]{1mm}{\multirow{2}{*}{\rotatebox[origin=c]{90}{{\scriptsize{\textsc{Divers.}}}}}} 
&\textsc{ConvexFall}
&40.4/12.8/36.3
&39.0/10.3/35.3
&17.3/3.2/14.2

\\
&\textsc{Heuristic}
&39.7/12.4/35.6
&38.1/9.8/34.5
&17.2/3.2/14.2
\\
\midrule
\parbox[t]{1mm}{\multirow{2}{*}{\rotatebox[origin=c]{90}{{\scriptsize{\textsc{Import.}}}}}}
&\textsc{N-Nearest}
&\textbf{41.4}/13.2/37.3
&\textbf{43.1/12.7/39.5}
&20.6/4.4/16.5
\\
&\textsc{K-Nearest}
&41.0/14.0/36.9
&40.0/12.3/36.6
&15.1/3.6/12.3
\\
\bottomrule
\end{tabularx}

\vspace{5mm}

\begin{tabularx}{.99\linewidth}{@{}c@{\hskip 1mm}|@{\hskip 0mm}r@{\hskip 0mm}|
@{\hskip 3mm} c@{\hskip 1mm}c@{\hskip 1mm}c  
@{\hskip 3mm} c@{\hskip 1mm}c@{\hskip 1mm}c 
@{\hskip 3mm} c@{\hskip 1mm}c@{\hskip 1mm}c 
@{\hskip 3mm} c@{\hskip 1mm}c@{\hskip 1mm}c@{}}
\toprule
&&\multicolumn{1}{c}{\texttt{AMI}}
&\multicolumn{1}{c}{\texttt{BookSum}} 
&\multicolumn{1}{c}{\texttt{MScript}} \\
\midrule
&& R-1/2/L 
& R-1/2/L
& R-1/2/L \\
\midrule
&\textsc{Random} 
&17.4/2.2/16.3
&41.6/7.0/39.6
&12.2/0.7/11.3\\
&\textsc{Oracle} 
&42.8/12.3/40.9
&52.0/14.7/50.2
&33.5/7.3/31.7\\
\midrule
\parbox[t]{1mm}{\multirow{3}{*}{\rotatebox[origin=c]{90}{{\scriptsize{\textsc{Position}}}}}} & \textsc{First-k}
&16.4/2.3/15.5
&\textbf{40.8/7.6/38.9}
&10.3/\textbf{1.1}/9.4\\
&\textsc{Last-k}
&11.1/1.7/10.5
&37.6/5.8/36.1
&\textbf{13.4}/0.9/12.1\\
&\textsc{Middle-k}
&16.1/1.9/15.2
&39.4/6.6/37.7
&12.1/0.6/11.2\\
\midrule
\parbox[t]{1mm}{\multirow{2}{*}{\rotatebox[origin=c]{90}{{\scriptsize{\textsc{Divers.}}}}}} 
&\textsc{ConvexFall}
&\textbf{20.4/2.5/19.1}
&24.3/3.9/22.6
&12.8/0.7/11.9\\
&\textsc{Heuristic}
&15.7/1.5/15.0
&38.2/6.2/36.4
&9.7/0.5/9.1\\
\midrule
\parbox[t]{1mm}{\multirow{2}{*}{\rotatebox[origin=c]{90}{{\scriptsize{\textsc{Import.}}}}}}
&\textsc{N-Nearest}
&1.9/0.1/1.8
&39.3/6.9/37.4
&13.1/0.8/\textbf{12.2}\\
&\textsc{K-Nearest}
&0.0/0.0/0.0
&30.9/5.0/29.5
&1.0/0.0/1.0\\
\bottomrule
\end{tabularx}
\caption{\label{tab:analysis_multi_appendix} 
Full ROUGE-1/2/L F-Scores for different corpora w.r.t three sub-aspects algorithms.}
\end{table}

\begin{figure*}[ht!]
\vspace{0mm}
\centering
\subfloat[\footnotesize{CNNDM}]{
\includegraphics[trim=0.6cm 1.1cm 0.6cm 0.6cm,clip,width=.38\textwidth]{figs/triangle/triangle_sys_cnndm.pdf}}
\qquad
\subfloat[\footnotesize{XSum}]{
\includegraphics[trim=0.6cm 1.1cm 0.6cm 0.6cm,clip,width=.38\textwidth]{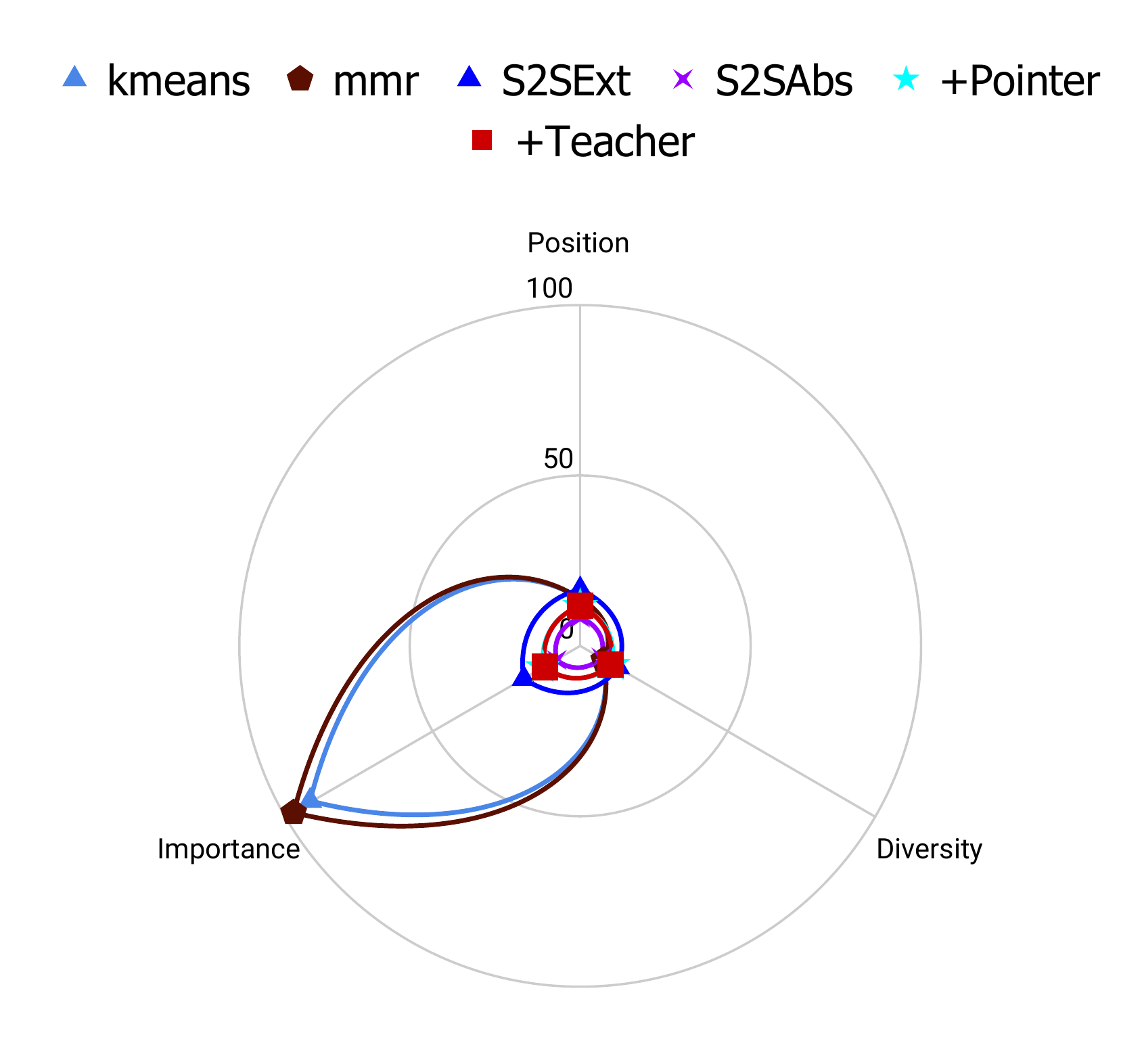}}
\\
\subfloat[\footnotesize{PeerRead}]{
\includegraphics[trim=0.6cm 1.1cm 0.6cm 0.6cm,clip,width=.38\textwidth]{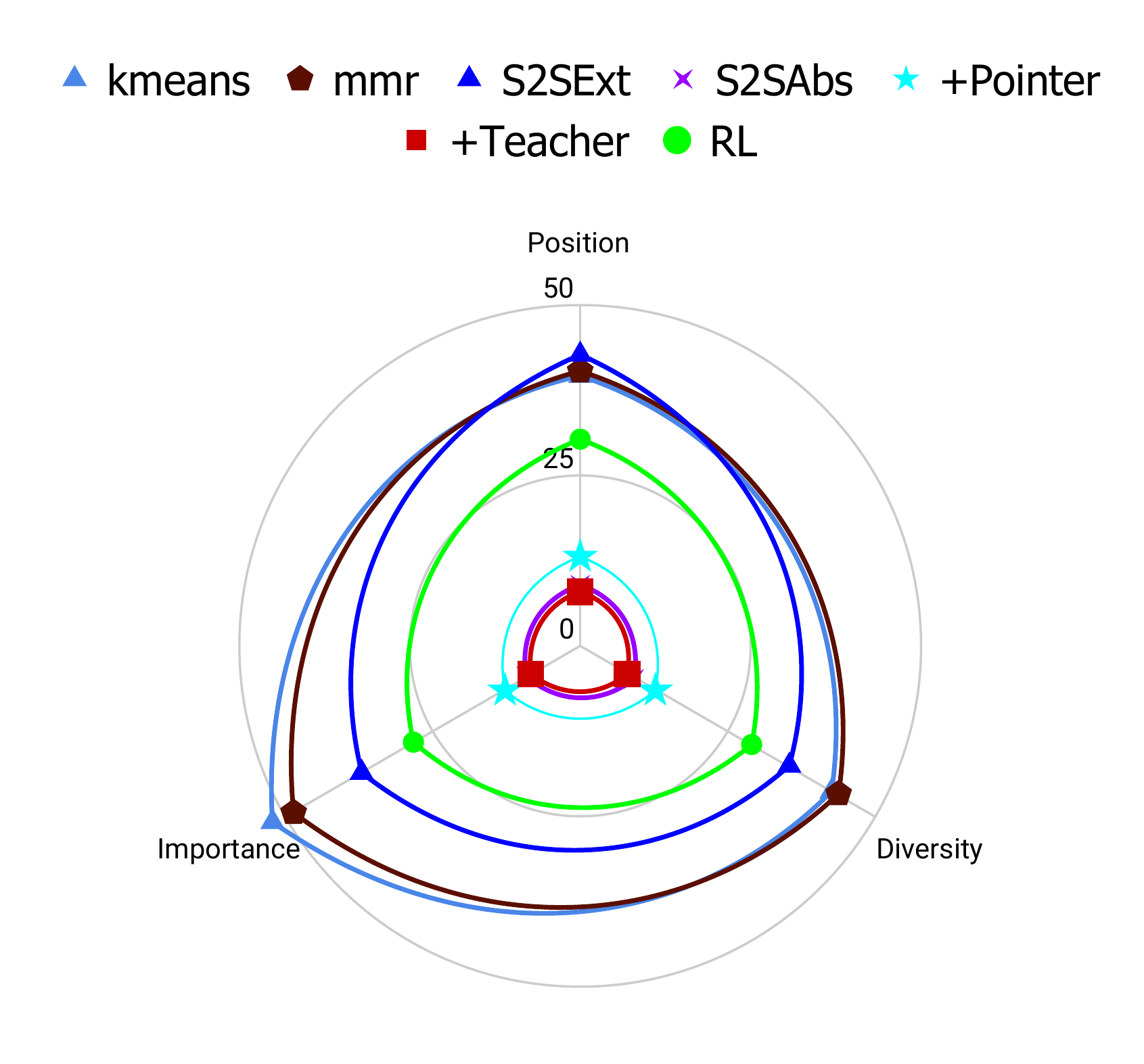}}
\qquad
\subfloat[\footnotesize{PubMed}]{
\includegraphics[trim=0.6cm 1.1cm 0.6cm 0.6cm,clip,width=.38\textwidth]{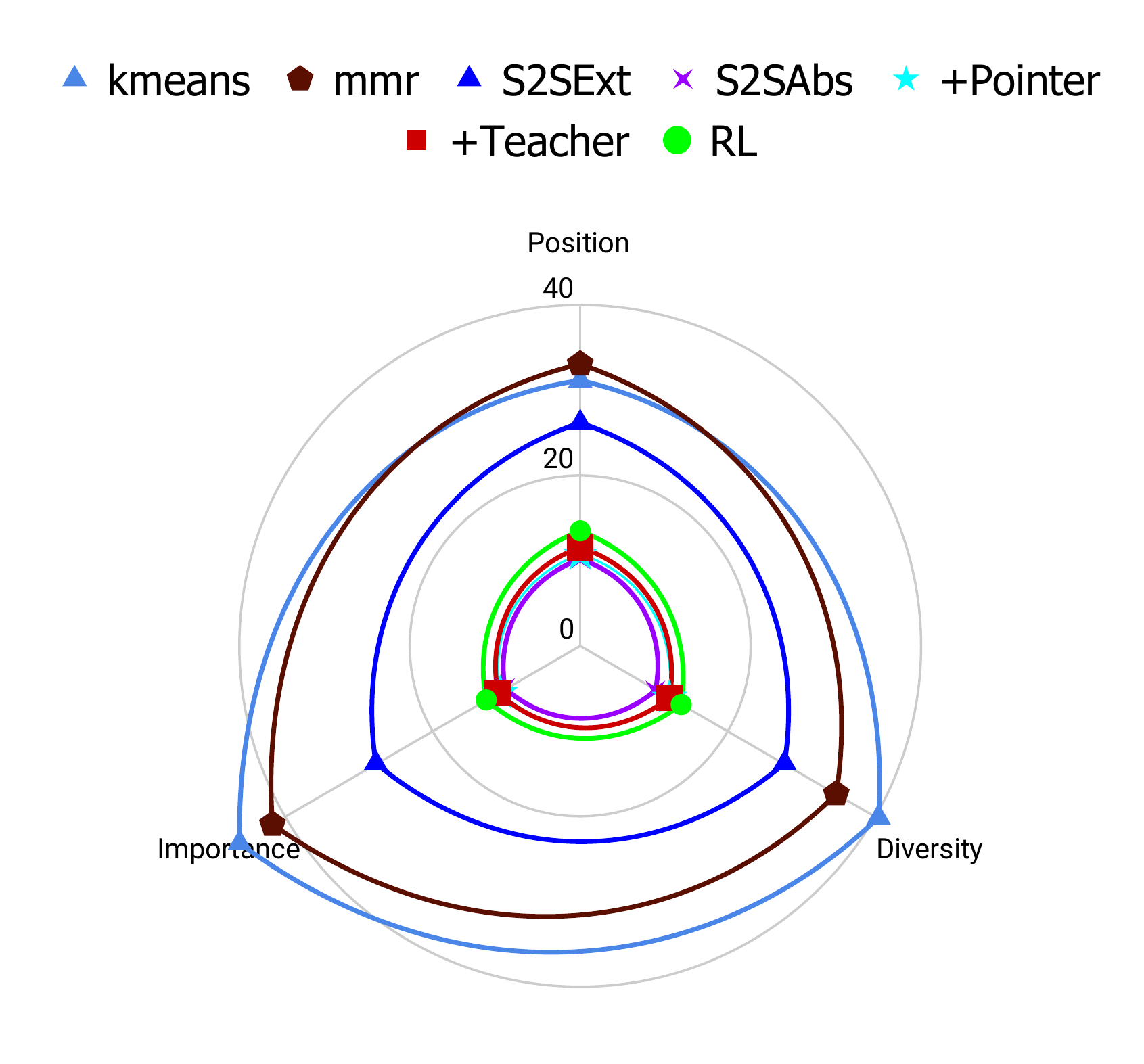}}
\\
\subfloat[\footnotesize{Reddit}]{
\includegraphics[trim=0.6cm 1.1cm 0.6cm 0.6cm,clip,width=.38\textwidth]{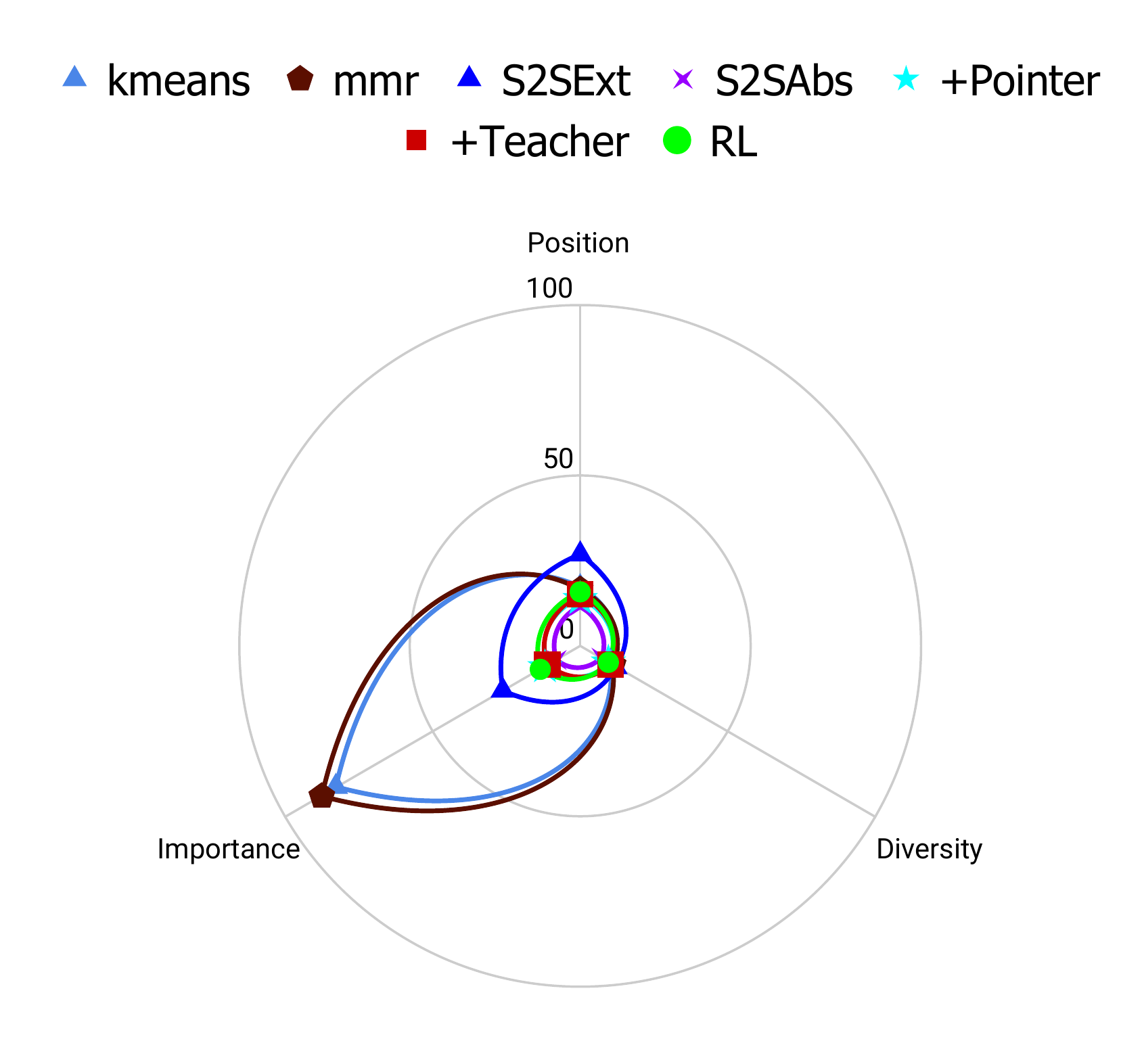}}
\qquad
\subfloat[\footnotesize{AMI}]{
\includegraphics[trim=0.6cm 1.1cm 0.6cm 0.6cm,clip,width=.38\textwidth]{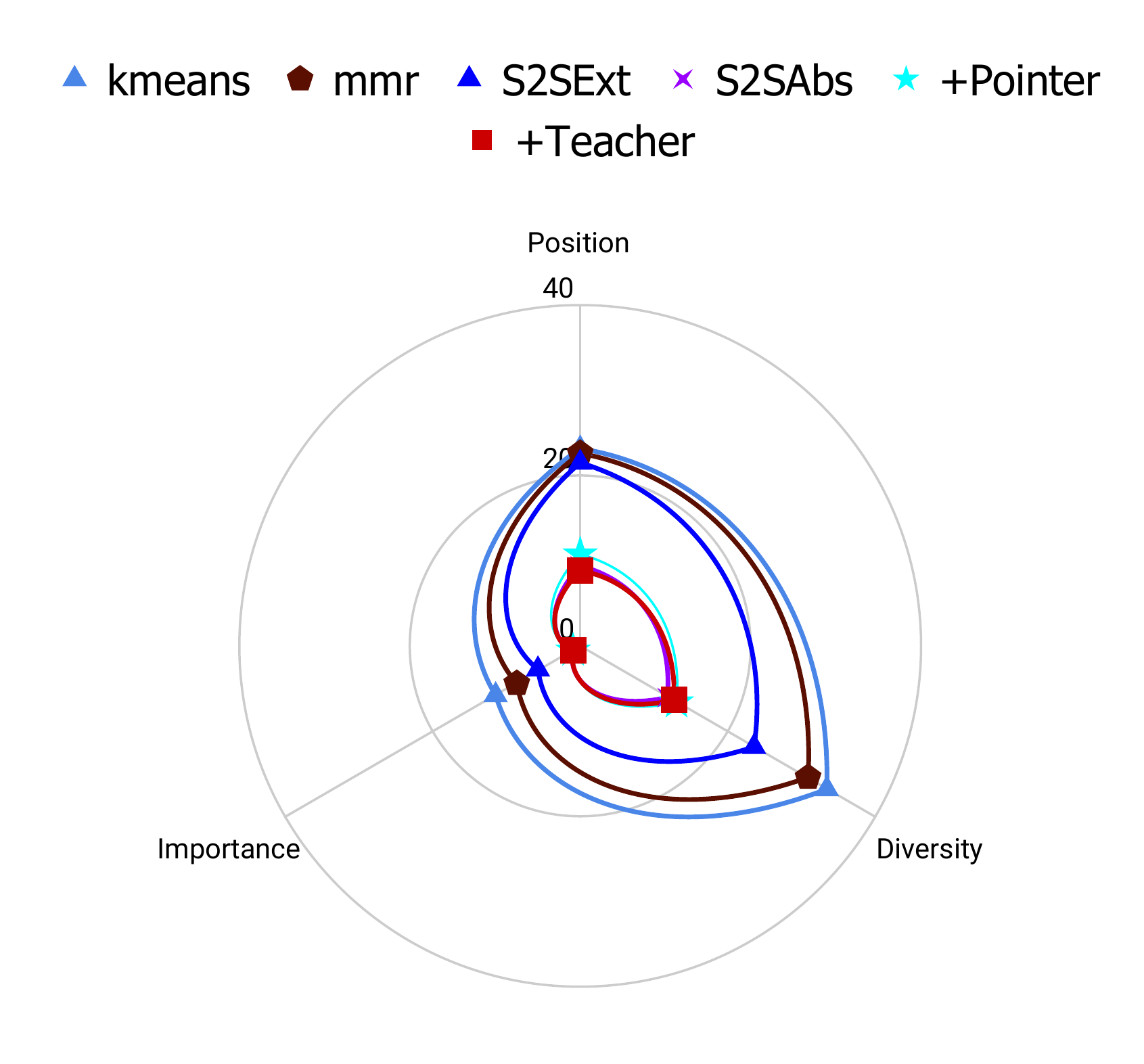}}
\\
\subfloat[\footnotesize{BookSum}]{
\includegraphics[trim=0.6cm 1.1cm 0.6cm 0.6cm,clip,width=.38\textwidth]{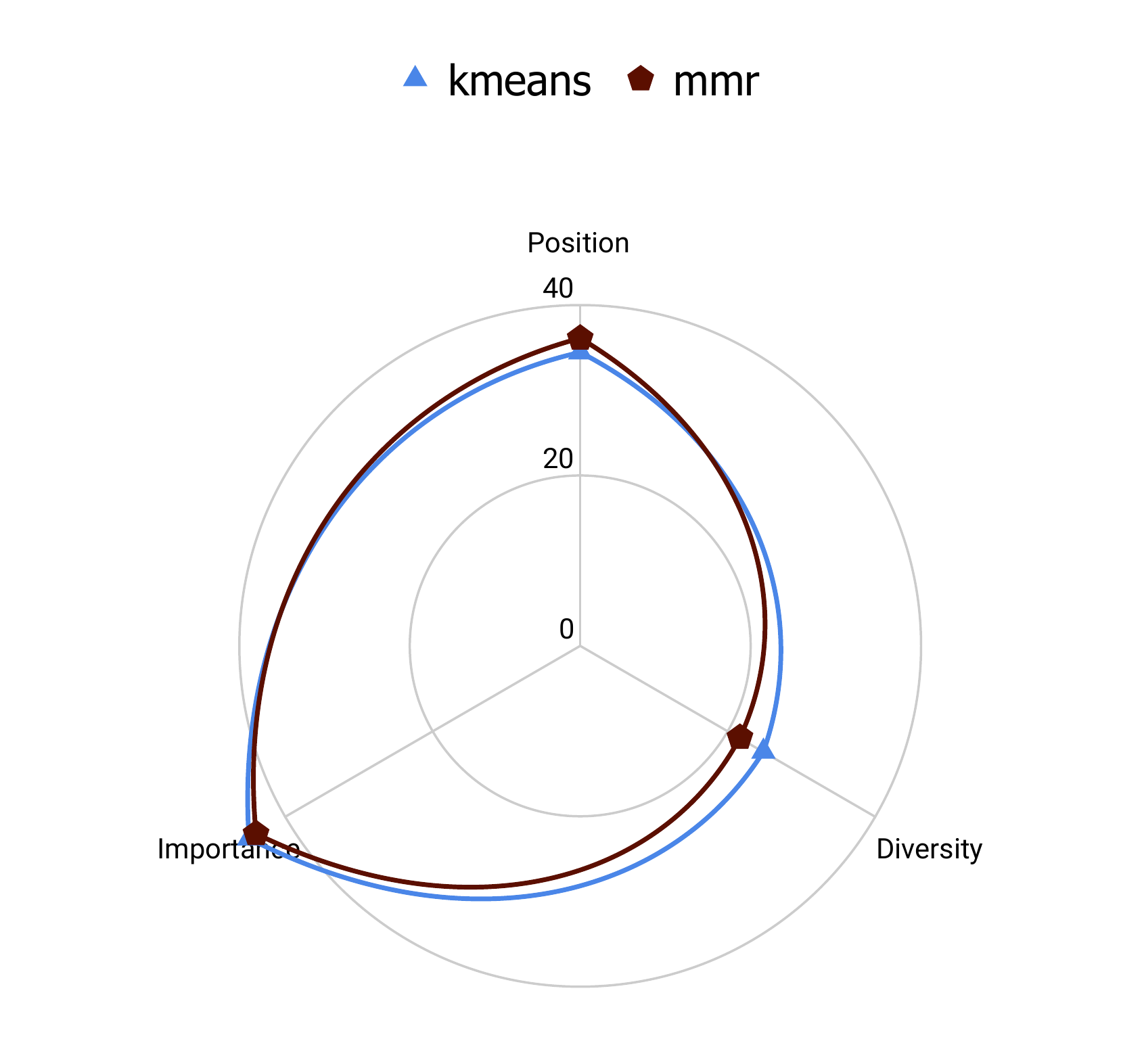}}
\qquad
\subfloat[\footnotesize{MScript}]{
\includegraphics[trim=0.6cm 1.1cm 0.6cm 0.6cm,clip,width=.38\textwidth]{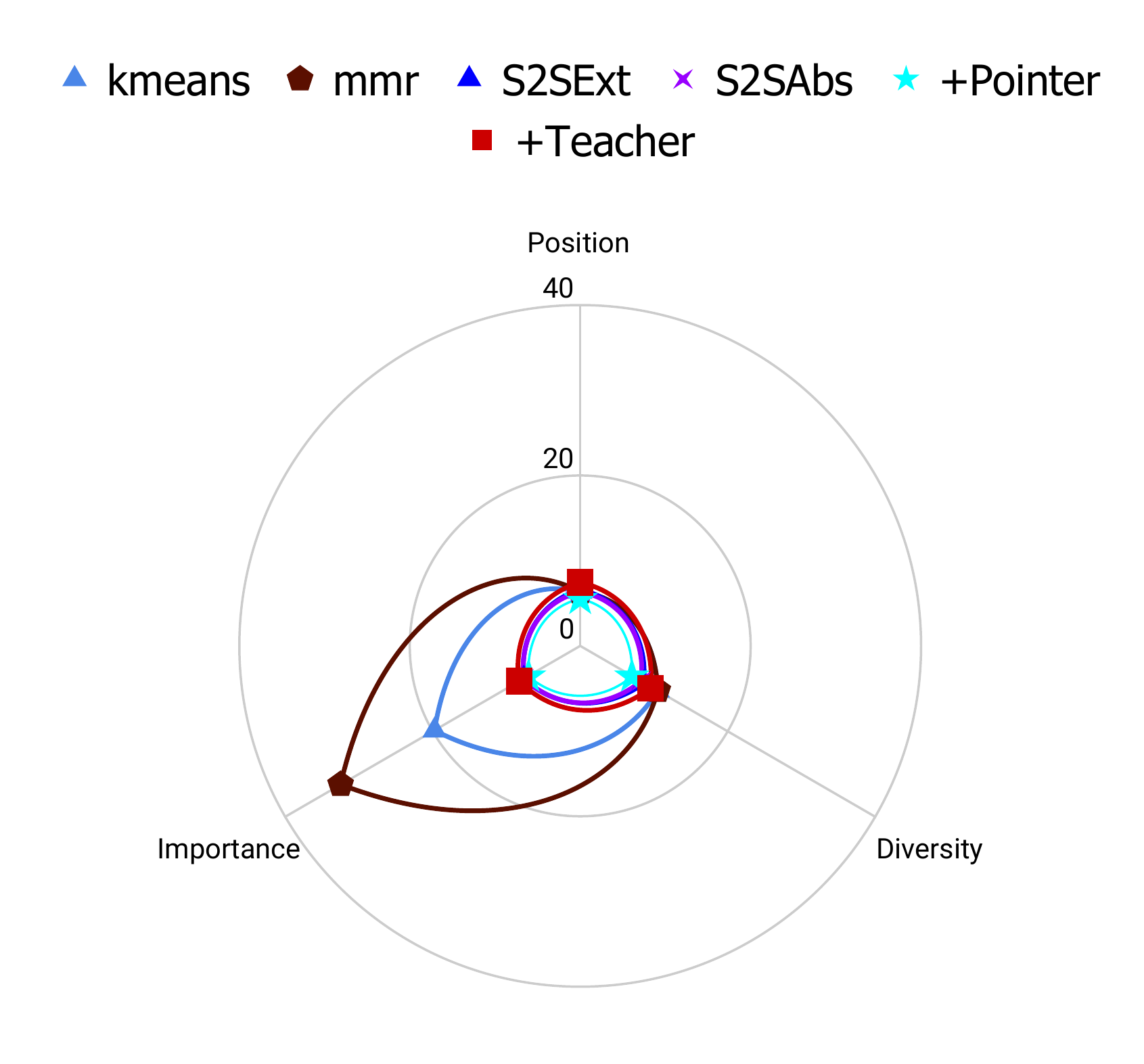}}
\caption{\label{fig:triangle_sys} System biases with the three sub-aspects per each corpus, showing what portion of aspect is used for each system.}
\end{figure*}

In Figure \ref{fig:triangle_sys}, we have more diagrams showing system biases toward each of three sub aspects.  We find that there exists a bias according to the corpus: for example in \texttt{Reddit}, many systems have a importance bias in common. On the other hand, systems are biased toward a diversity aspect in \texttt{AMI}. Also, some systems tend to be biased in certain aspect across the different corpus: systems such as \textit{KMeans} and \textit{MMR}, many corpora are biased toward a importance aspect. 






\end{appendix}

\end{document}